\documentclass[runningheads]{llncs}

\usepackage{eccv}

\usepackage{eccvabbrv}

\usepackage{graphicx}
\usepackage{import} 
\usepackage{booktabs}
\usepackage{float} 
\usepackage[linesnumbered,ruled,vlined]{algorithm2e}
\usepackage{multirow, multicol, booktabs}
\usepackage{enumitem}
\usepackage{tabularx}
\usepackage{array}
\newcolumntype{Y}{>{\centering\arraybackslash}X}
\usepackage{booktabs}
\usepackage{makecell}
\usepackage{wrapfig}
\usepackage[table]{xcolor}

\newcommand*{\methodname}[0]{\textsc{DiverseVAR}}
\newcommand{\sos}{\texttt{<SOS>}}

\SetAlCapHSkip{0pt}
\SetKwInput{KwInput}{Inputs}
\SetKwInput{KwOutput}{Outputs}
\SetKwInOut{KwParam}{Params}
\SetKwProg{myalg}{def}{}{}
\SetKwFunction{algo}{MultiScaleEnc}
\SetKwFunction{algost}{ScaleTravel}

\definecolor{darkred}{rgb}{0.7,0.2,0.1}
\definecolor{darkgreen}{rgb}{0,0.7,0}
\definecolor{orange}{RGB}{255,127,0}
\definecolor{ourpurple}{RGB}{127,127,204}
\definecolor{palgreen}{RGB}{51,179,179}
\definecolor{magenta}{RGB}{199,21,133}

\definecolor{MyGreen}{rgb}{0,0.7,0}
\definecolor{MyYellow}{rgb}{0.87,0.87,0}

\usepackage{dsfont}
\newcommand{\mathbbm}[1]{\ensuremath{\mathds{#1}}}

\usepackage[accsupp]{axessibility}  

\usepackage[hidelinks]{hyperref}

\usepackage{orcidlink}

\begin{document}

\addtocontents{toc}{\protect\setcounter{tocdepth}{-1}}

\title{DiverseVAR: Balancing Diversity and Quality of Next-Scale Visual Autoregressive Models}
\titlerunning{Balancing Diversity and Quality of Next-Scale Visual Autoregressive Models}


\author{%
    Mingue Park\textsuperscript{*}\orcidlink{0009-0003-3973-9742} \and
    Prin Phunyaphibarn\textsuperscript{*}\orcidlink{0009-0002-2821-6666} \and
    Phillip Y. Lee\orcidlink{0009-0000-7614-261X} \and \\
    Minhyuk Sung\orcidlink{0000-0001-7428-9570}
}

\authorrunning{Park et al.}

\institute{%
    KAIST, Daejeon, South Korea\\
    \email{\{kicikicik, prin10517, phillip0701, mhsung\}@kaist.ac.kr}\\
}

\maketitle
\begin{center}
    \centering
    \captionsetup{type=figure}
    \includegraphics[width=0.92\textwidth]{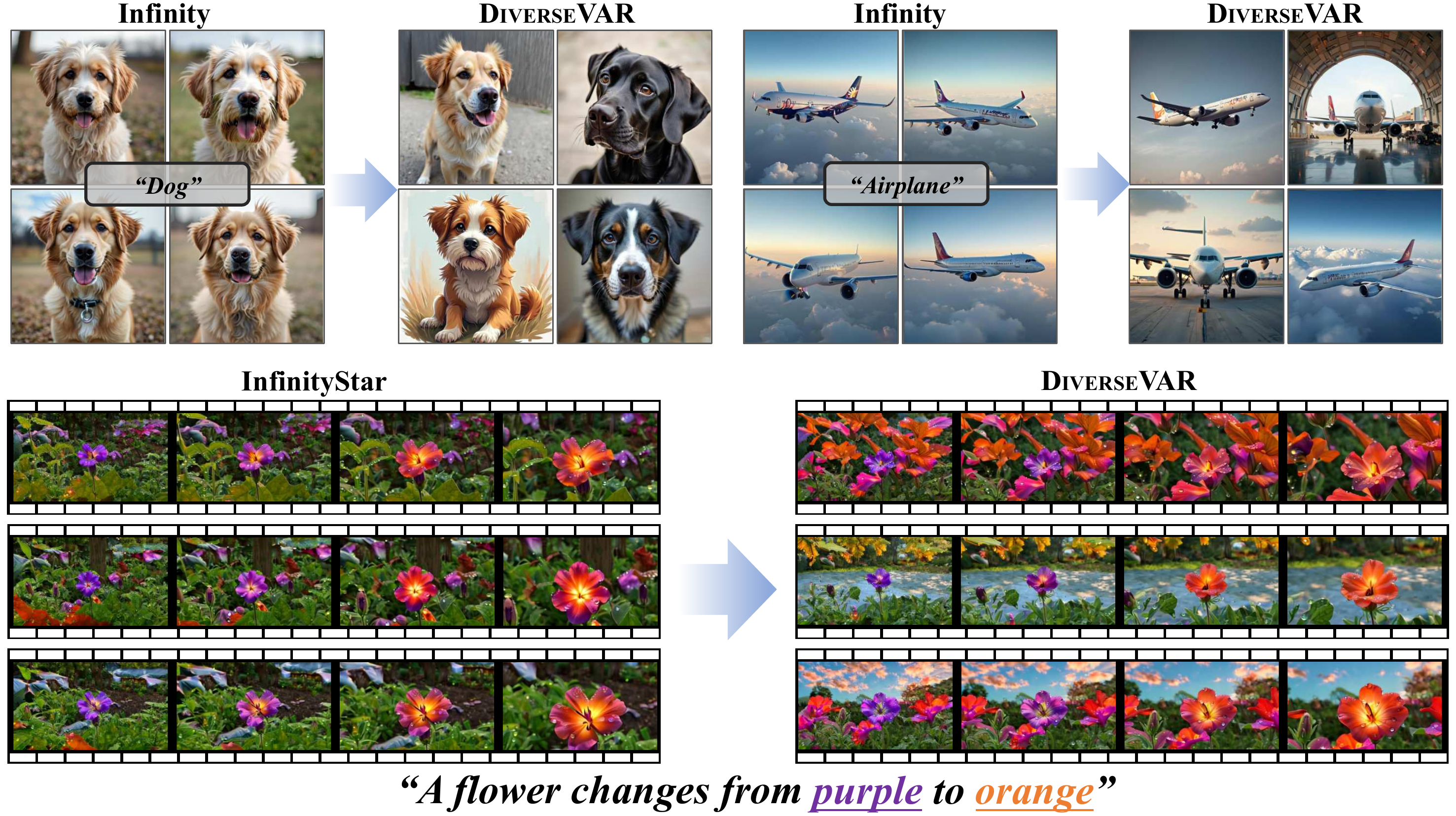}
    \vspace{-0.5\baselineskip}
    \caption{\textbf{\methodname{}.} Text-conditioned VAR models (Infinity~\cite{han2024infinity} and InfinityStar~\cite{liu2025:infinitystar}) exhibit a clear lack of per-prompt diversity in image and video generation. \methodname{} combines diversity-enhancing annealing strategies with a novel latent refinement technique to increase the diversity while maintaining visual quality.}
    \label{fig:teaser}
    \vspace{-0.5\baselineskip}
\end{center}

\begingroup
\renewcommand{\thefootnote}{*}
\footnotetext{Equal contribution.}
\endgroup

\begin{abstract}
We introduce \methodname{}, a test-time framework that enhances the output diversity of text-conditioned visual autoregressive models (VAR) without additional training or substantial computational overhead.
While VAR models have recently emerged as strong competitors of diffusion and flow models for image and video generation, they suffer from a critical diversity limitation: even simple prompts often produce nearly identical outputs. This issue has largely gone unnoticed amid the field's predominant focus on image quality.
We address this limitation in two stages. First, we systematically explore diversity enhancement techniques by injecting noise into different components of VAR at test time, finding that noise injection into the text embedding yields the best diversity gains. However, this comes at the cost of sharp degradation in image quality. To recover quality, we propose \textbf{scale-travel}: a latent refinement technique that leverages a multi-scale autoencoder to extract coarse-scale tokens to resume generation from intermediate stages. Extensive experiments demonstrate that combining text-embedding noise injection with scale-travel refinement substantially improves diversity while minimizing quality degradation, advancing the diversity-quality Pareto front. Project page: \url{https://diverse-var.github.io/}
\vspace{-0.5\baselineskip}
\keywords{Visual Autoregressive Models \and Diversity}
\end{abstract}

\vspace{-0.5\baselineskip}
\section{Introduction}
\label{sec:intro}
\vspace{-0.5\baselineskip}

Visual autoregressive (VAR) models~\cite{tian2024visual,han2024infinity, jiao2025flexvar, tang2025hart, zhuang2025vargpt, li2024controlvar, zhang2024varclip} have recently emerged as strong competitors to diffusion and flow models for image and video generation, achieving comparable output quality. 
The recent breakthrough in VAR lies in shifting the prediction paradigm from next-token prediction to \emph{next-scale} prediction~\cite{tian2024visual}. VAR represents an image as a pyramid of latent patches, structured from the lowest resolution (a single coarse patch) to progressively higher resolutions; within each level, VAR generates patches jointly, while across levels, finer-scale patches are generated autoregressively, conditioned on the previously generated coarser level. This shift to autoregressive next-scale prediction allowed causal autoregressive models to outperform diffusion models for the first time, while also being significantly faster than both diffusion models and raster-order autoregressive models. Recent works~\cite{guo2025fastvar, li2025skipvar} have further improved the speed of VAR, allowing for fast and high-quality image generation.

Recent works have developed large-scale VARs for text-conditioned image and video generation.
However, text-conditioned VARs still lack diversity, often generating similar outputs when given the same prompt (Fig.~\ref{fig:teaser}). This lack of diversity has also been observed in other generative models such as diffusion models~\cite{sadat2023cads, somepalli2023understanding} and GANs~\cite{liu2020diversegan}. While various techniques such as CFG annealing~\cite{Kynkaanniemi2024cfginterval} and CADS~\cite{sadat2023cads} have been proposed to improve diversity in diffusion models, improving diversity in text-conditioned VAR generation remains an open area of research. In this work, we address this issue in text-conditioned VARs through inference-time modifications, without requiring model retraining, fine-tuning, or substantial additional compute at generation.

Inspired by diversity-enhancement techniques in diffusion models~\cite{sadat2023cads, Kynkaanniemi2024cfginterval}, we first explore their impact in the scale-wise autoregressive setting of VAR. These diversity-enhancement methods mainly fall into two categories. The first is CFG weight scheduling~\cite{ho2021classifierfree, Kynkaanniemi2024cfginterval}: similar scheduling can be applied across scales during the autoregressive next-scale generation in VAR. The second technique is noise injection into the text embedding (also known as condition-annealing~\cite{sadat2023cads}), which can likewise be applied to VAR models. Both approaches involve a trade-off between diversity and quality; increased diversity typically leads to reduced quality. Empirically, we observe that the second approach (condition-annealing) is more effective for VAR models but results in reduced image quality.

To further mitigate the degradation in output quality, we propose a novel latent refinement framework called \textbf{Scale-Travel}. 
Motivated by image editing~\cite{meng2021sdedit}, inpainting~\cite{lugmayr2022repaint}, and reward alignment techniques~\cite{ma2025inference}, which perform latent refinement by briefly reverting part of the generative trajectory before resuming generation, we explore a novel mechanism for VAR along the \emph{scale axis}, which is non-trivial in the autoregressive generation process. Our method exploits the latent multi-scale autoencoder to enable an approximate scale rollback and refinement during next-scale autoregressive generation. Specifically, given a partial sequence of token scales, we encode their accumulation into a full sequence of token scales, and then retain only the first few scales. In this way, we can refine the partial sequence while effectively reverting to a previous scale by discarding the newly encoded finer-scale tokens. Incorporating this VAR-based ``Scale-Travel'' module into the autoregressive generation process significantly enhances realism, helping to overcome the diversity-realism trade-off caused by condition-annealing.

We empirically show that \methodname{} yields the best diversity-quality tradeoff on both text-conditioned image and video generation and is robust across CFG scales and varying temperatures and nucleus sampling~\cite{holtzman2020:nucleus} parameters.
\vspace{-0.5\baselineskip}
\section{Related Work}
\label{sec:related}

\subsection{Autoregressive Visual Generation}
\label{subsec:related_autoregressive_image_generation}
Building on the success of transformer-based autoregressive (AR) modeling in LLMs~\cite{vaswani2023attention, brown2020gpt3}, researchers have increasingly explored similar techniques for image and video generation. 
In the image domain, AR methods differ primarily by the order in which they process local patches (or tokens). First, \textbf{(1) raster-scan order AR} methods take an analogous approach to language modeling by flattening and quantizing a 2D image into a 1D sequence of discrete tokens, and then training an AR model for next-token prediction~\cite{van2016pixelrnn, van2016pixelcnn, salimans2017pixelcnn++, reed2017parallel, chen2020igpt, esser2021vqgan, ramesh2021dalle, razavi2019vqvae2, lee2022rqtransformer, yu2022parti, sun2024llamagen, fan2024fluid, wang2025bridging}. Early works like PixelCNN~\cite{van2016pixelcnn} perform autoregressive generation in the pixel space, while more recent works, including VQGAN~\cite{esser2021vqgan} and LlamaGen~\cite{sun2024llamagen}, perform it in the latent space~\cite{van2017vqvae}. Follow-up works propose alternative token ordering schemes to address the inflexibility of a fixed raster-scan order~\cite{chen2018pixelsnail, pang2024randar}. 
In the video domain, transformer-based AR models have also been actively explored, extending next-token prediction to spatiotemporal token sequences~\cite{deng2025:nova, wang2024:emu3, kondratyuk2024:videopoet}.
Going further, \textbf{(2) masked AR} methods remove the strict unidirectional raster-scan by randomly masking subsets of tokens and predicting them in parallel with full attention over the tokens~\cite{chang2022maskgit, yu2023magvit, luo2024openmagvit2, chang2023muse, lezama2022tokencritic, ni2024revisiting, hur2024unlocking, li2024autoregressive, bai2025meissonic, besnier2025halton, you2025effective, xie2025showo, weber2024maskbit}. Represented by MaskGIT~\cite{chang2022maskgit}, this parallel mask prediction introduces significant inference acceleration and also provides a connection between autoregressive modeling and discrete diffusion~\cite{gat2024discrete, austin2021structured}, as demonstrated by Xie \textit{et al.}~\cite{xie2025showo}.

Unlike the AR methods above, which predict individual tokens, visual autoregressive (VAR) model~\cite{tian2024visual} represents an image as a sequence of multi-scale token maps. A transformer is trained to progressively predict each next higher-resolution token map. \textbf{(3) VAR} better respects the 2D grid structure of images, scales effectively, and outperforms both raster-scan order AR~\cite{lee2022rqtransformer, esser2021vqgan}, masked AR~\cite{chang2022maskgit}, along with diffusion-based methods~\cite{dhariwal2021adm, peebles2023dit} on ImageNet~\cite{deng2009imagenet}. 
Recent VAR-based models, such as Infinity~\cite{han2024infinity} and Switti~\cite{voronov2024switti} for images, as well as InfinityStar~\cite{liu2025:infinitystar} for text-conditioned video generation, demonstrate that VAR scales effectively to large datasets. Ongoing work continues to introduce VAR variants~\cite{ren2024mvar, jiao2025flexvar, yu2025frequency, zhang2024varclip, chen2024collaborative, guo2025fastvar, yu2024rar, huang2025nfig} and their adaptation to downstream applications like editing~\cite{wang2025training, wangeditinfinity}, personalization~\cite{chung2025fine}, controllable generation~\cite{park2025training, yao2024car, li2024controlvar}, and style transfer~\cite{nguyen2025csd}. Other works~\cite{guo2025fastvar, li2025memory, li2025skipvar} improve the efficiency in VAR-based models, significantly decreasing memory cost and runtime.

\subsection{Diversity Enhancement for Visual Generation}
\label{subsec:related_diversity}
Enhancing diversity is a crucial task in visual generation for both diffusion and autoregressive (AR) models. 
Originally introduced in the context of diffusion models, CFG (Classifier-Free Guidance) sampling~\cite{ho2021classifierfree} is now widely used in both domains.
Specifically in AR models, diversity has been promoted by sampling tokens probabilistically~\cite{teng2024accelerating, sun2024llamagen}. 
However, the CFG technique tends to narrow the categorical token distribution, improving image quality but simultaneously diminishing output diversity. To mitigate mode collapse, dynamic CFG scheduling~\cite{Kynkaanniemi2024cfginterval, wang2024analysis} has proven effective for diffusion models. 
Another diversity enhancement technique for diffusion models is CADS~\cite{sadat2023cads}, which demonstrated that injecting random Gaussian noise into the text embedding allows control over the quality–diversity trade-off during inference.
To the best of our knowledge, we are the first to systematically increase the diversity of VAR models. In this work, we extend these approaches to VAR~\cite{han2024infinity, tian2024visual} and investigate which components play a key role in enhancing diversity. Additionally, we introduce our novel method, \methodname{}, a method designed to address the identified challenges. 

\vspace{-0.5\baselineskip}
\section{Background: Next-Scale Visual Autoregressive Models}
\label{sec:background}
\vspace{-0.5\baselineskip}
In this section, we provide background on next-scale visual autoregressive (VAR) models, outlining their key principles and architectural components. We then explain how conditional signals are integrated into VAR and discuss a critical drawback, namely the limited diversity of the generated visual content.

\vspace{-0.5\baselineskip}
\paragraph{VAR: Next-Scale Prediction.}
Although autoregressive (AR) models extend the language-modeling paradigm to image generation, the standard next-token objective suffers from two major flaws. First, flattening an image or video into a 1D token sequence retains bidirectional dependencies, violating the unidirectional (\ie, causal) assumption in AR sampling. Second, the flattened representation fails to exploit the inherent spatial locality of an image. To resolve these issues, Tian \etal~\cite{tian2024visual} introduced VAR---\emph{Visual Autoregressive Model}---which substitutes next-token prediction with a \emph{next-scale} objective. 
Given an image $I$, a feature map $\mathbf{Z}\in \mathbb{R}^{H\times W\times D}$ is obtained through an encoder (\eg, VQ-VAE~\cite{van2017vqvae}). Rather than patchifying the feature map, VAR employs \emph{multi-scale encoding} in which $\mathbf{Z}$ is decomposed into $K$ \textbf{residual token maps} $(\mathbf{r}_1, \cdots, \mathbf{r}_K)$ of sizes $\mathbf{s}_k=(h_k, w_k)$. At stage $k$ the model constructs a \textbf{feature map} $\mathbf{Z}_k$ by upsampling all preceding token maps to the final resolution $\mathbf{s}_K=(H, W)$ and aggregating them:

\vspace{-0.5\baselineskip}
\begin{equation}
    \label{eq:var-canvas}
    \mathbf{Z}_k = \sum_{i=1}^k \texttt{up}(\mathbf{r}_i, \mathbf{s}_K),
\end{equation}
where $\texttt{up}(\cdot, \cdot)$ denotes bilinear upsampling. The joint probability of the token maps then factorizes over \emph{scales} rather than token orders:
\vspace{-0.5\baselineskip}
\begin{align}
    p(\mathbf{r}_1, \mathbf{r}_2, ..., \mathbf{r}_K) = \prod^{K}_{k=1} p(\mathbf{r}_k | \mathbf{r}_1, \mathbf{r}_2, ..., \mathbf{r}_{k-1}).
    \label{eq:next_scale_prediction}
\end{align}
with $\mathbf{r}_k \in \mathbb{R}^{h_k \times w_k \times D}$. This next-scale objective preserves spatial locality while preserving the unidirectional assumption for autoregressive models.

\vspace{-0.5\baselineskip}
\paragraph{Conditional Generation with VAR.}
The high scalability of VAR has led to its rapid adoption in text-conditioned image generation~\cite{han2024infinity, tang2025hart, zhuang2025vargpt, zhuang2025vargptv11}. Text condition is incorporated into VAR through two complementary mechanisms: \textbf{(1) cross-attention with text embeddings}, and \textbf{(2) \sos~token initialization}. In the former, the text embedding is first projected to obtain a condition embedding $\mathbf{c}$. $\mathbf{c}$ is then passed through projection networks to obtain key and value matrices which cross-attends with the image tokens, similar to cross-attention in diffusion models like Stable Diffusion~\cite{rombach2022stablediffusion}. For the latter, the text embedding is projected to produce the \sos---\emph{Start-of-Sentence}---token $\mathbf{s}$ which initializes the generation as done in typical autoregressive models.

\vspace{-0.5\baselineskip}
\paragraph{Lack of Diversity in VAR.}
In this work, we recognize a significant lack of diversity in text-conditioned VAR models. While VAR models make use of basic techniques such as temperature annealing and linear CFG scheduling to increase diversity, we find that these methods provide limited improvements. Motivated by this observation, we explore and propose new diversity-enhancement strategies to boost their diversity while maintaining the image quality and conditional fidelity in VAR models. 
Beyond generating high-quality images that match a given text prompt, a critical desideratum for text-conditioned image generation is to be able to provide \emph{diverse} outputs from a single condition (\ie, text prompt). This diversity not only offers users a range of options to choose from based on their preferences, but also acts as a prerequisite for test-time search algorithms~\cite{xie2025sana, kim2025inference} for downstream applications. Although recent VAR-based models such as Infinity~\cite{han2024infinity} achieve impressive quality, they suffer from severely limited diversity, as demonstrated in Fig.~\ref{fig:teaser}. In this work we adopt Sadat~\etal~\cite{sadat2023cads} definition of diversity, which refers to the model's ability to generate varied outputs for a fixed condition $\mathbf{c}$ under different random seeds.

\begin{figure*}[t!]
    \centering
    \includegraphics[width=\textwidth]{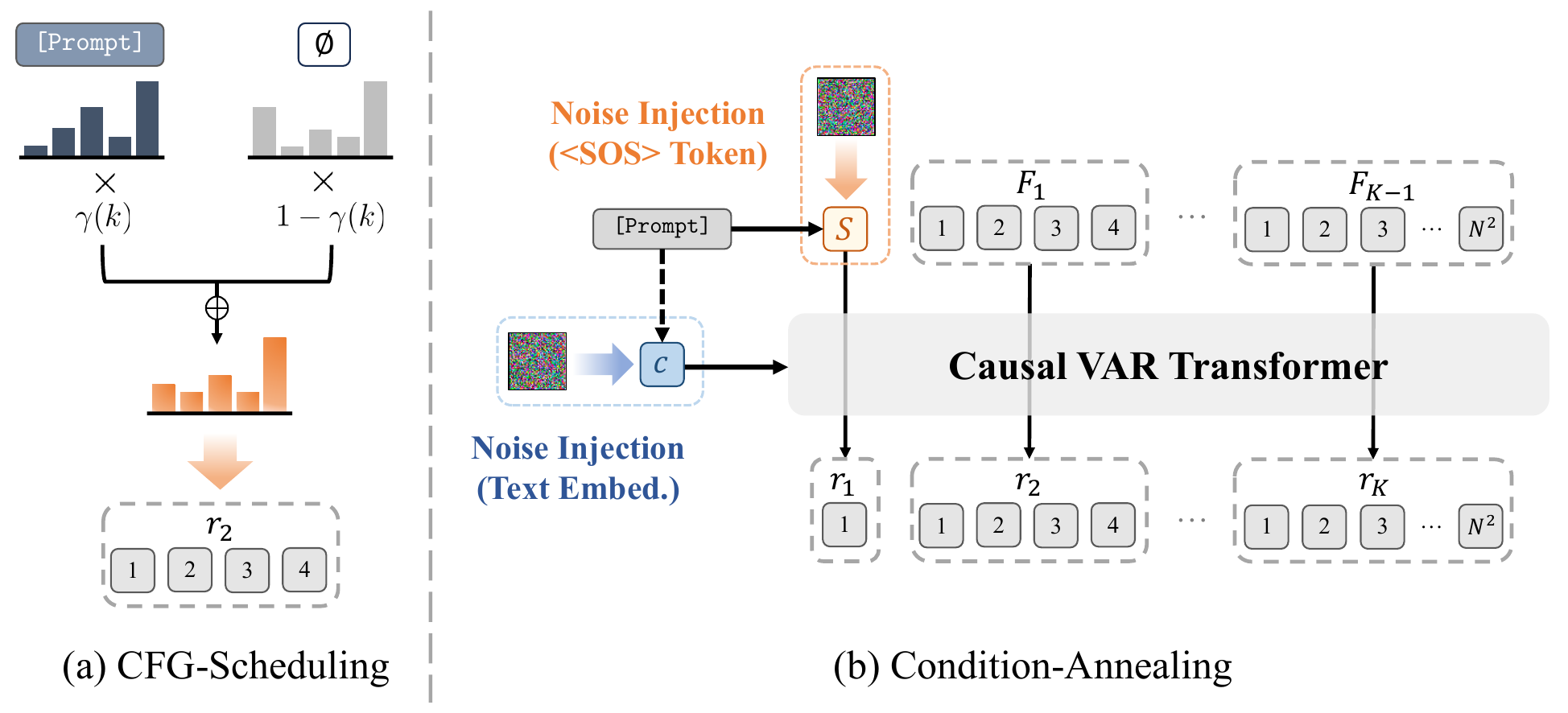}
    \vspace{-\baselineskip}
    \caption{\textbf{Diversity-Enhancement Techniques.} We explore two main options for diversity enhancement in VAR: (a) CFG-scheduling and (b) condition-annealing. CFG-scheduling modulates the CFG scale over sampling steps to mitigate mode collapse. Condition-annealing injects noise into the text-embedding or the \sos~token.}
    \label{fig:condition_annealing_comparison}
\end{figure*}

\section{\methodname{}}
\label{sec:method}
\vspace{-0.5\baselineskip}

We present \methodname{}, a training-free framework for enhancing diversity in text-conditioned VAR models. First, we explore inference-time techniques designed for diversity enhancements and compare their impact on diversity and quality in VAR (Sec.~\ref{subsec:method_condition_annealing}). 
Observing a critical trade-off between these metrics, we then propose a simple latent refinement strategy by taking advantage of the multi-scale structure of VAR (Sec.~\ref{subsec:method_scale-travel}).
Finally, we combine these two stages into a unified diversify-then-refine pipeline, achieving significant diversity gains with minimal quality degradation.

\subsection{Exploring Diversity-Enhancement Techniques}
\label{subsec:method_condition_annealing}
In this section, we search for the best diversity-enhancement strategy for VAR models.
Observing that VAR produces nearly identical samples for the same text prompt due to the condition signal dominating the generation process, we first explore two inference-time techniques popularly used to mitigate mode collapse as shown in Fig.~\ref{fig:condition_annealing_comparison}: (1) custom classifier-free guidance (CFG) schedules~\cite{ho2021classifierfree}, and (2) condition-annealing by injecting noise into the condition embedding~\cite{sadat2023cads}.

\paragraph{Option 1: CFG Scheduling.}
CFG~\cite{ho2021classifierfree} modulates the influence of the condition in diffusion and VAR~\cite{tian2024visual, han2024infinity}. Formally, CFG adjusts the model's prediction via
\begin{align}
    \mathbf{Q}_\text{CFG} = (1+\omega)\mathbf{Q}_c - \omega \mathbf{Q}_\emptyset,
\end{align}
where $\mathbf{Q}_\mathbf{c}$ is the conditional prediction (\ie, score estimates in diffusion and logit outputs in VAR) given a condition $\mathbf{c}$, $\mathbf{Q}_\emptyset$ is the unconditional prediction, and $\omega$ is the guidance weight.
Building on the trend that higher $\omega$ leads to higher condition alignment while sacrificing diversity, Kynk{\"a}{\"a}nniemi \etal~\cite{Kynkaanniemi2024cfginterval} show that carefully scheduling the guidance weight $\omega$ throughout generation can boost diversity while retaining image quality. While VAR models have employed \emph{linear} CFG scheduling, there has been no systematic study on its impact on diversity nor of the effects of other CFG schedules. We evaluate multiple CFG schedules $\omega(k)$ over the scale index $k \in [1, K]$ for VAR, analogous to timestep $t\in [0,T]$ in diffusion models.
\begin{itemize}[leftmargin=8pt]
    \item \underline{Piecewise Constant}~\cite{Kynkaanniemi2024cfginterval}: \colorbox{gray!10}{\small $\omega(k) = \mathbbm{1}[k \in \mathcal{K}] \omega_1 + \mathbbm{1}[k \not\in \mathcal{K}] \omega_K$}\\ where $\mathcal{K} := \{ k_{\text{min}},\dots, k_{\text{max}} \}$, and $\omega_1$ and $\omega_K$ are tunable hyperparameters. 
    \item \underline{Interpolation}~\cite{wang2024analysis}: \colorbox{gray!10}{\small $\omega(k) = (1 - \gamma(k)) \omega_{1} + \gamma(k) \omega_K$} \\ Here, $\gamma(1)=0$ and $\gamma(\ell) = 1$ for all $\ell \ge k_{\max}$, and we test several variants of the scheduling function $\gamma(\cdot)$ suggested by Wang~\etal~\cite{wang2024analysis}. $\omega_1$ and $\omega_K$ denote the initial and final guidance weight, respectively. Details on each scheduler and ablations for $k_{\text{max}}$ are provided in the \textbf{supplementary}.
\end{itemize}

\paragraph{Option 2: Condition-Annealing via Noise Injection.}
Sadat~\etal~\cite{sadat2023cads} introduce condition-annealing diffusion sampler (CADS), a technique that perturbs the condition by injecting noise into the text embedding, starting with full noise at the initial generation steps. This simple modification has proven effective for diversity in diffusion models as it weakens the condition's dominance in the sampling process. We explore condition-annealing in VAR on its two primary condition sources: (1) text embedding and (2) \sos~token. 

\begin{itemize}[leftmargin=8pt]
    
    \item \underline{Text Embedding}: \colorbox{gray!10}{\small $\mathbf{\hat{c}}=\sqrt{1 - \alpha(k)} \mathbf{c} + \sqrt{\alpha(k)} \epsilon,\; \epsilon \sim \mathcal{N}(0,\mathbf{I})$} \\ Following Sadat~\etal~\cite{sadat2023cads}, we add noise $\epsilon$ to the text embedding \textbf{c}, modulated by an annealing schedule $\alpha(\cdot)$. We explore different choices for $\alpha(\cdot)$ in the \textbf{supplementary}.
    
    \item \underline{\sos~Token}: \colorbox{gray!10}{\small $\hat{\mathbf{s}} = \sqrt{1 - \beta(k)} \mathbf{s} + \sqrt{\beta(k)} \epsilon, \; \epsilon \sim \mathcal{N}(0,\mathbf{I})$} \\ Since autoregressive models prepend a special \sos~token $\mathbf{s}$ to initialize the generation process, we also test injecting noise into $\mathbf{s}$ according to schedule $\beta(\cdot)$.
    
\end{itemize}

\paragraph{Empirical Analysis.} We compare the diversity-quality tradeoff for each of the aforementioned options, and find that injecting noise into the text-embedding yields the best results, as shown in Fig.~\ref{fig:pareto}. 
Although the top-performing configuration does boost diversity, the outputs often exhibit noticeable degradations and visual artifacts (see Fig.~\ref{fig:teaser} and Fig.~\ref{fig:main_qualitative}). We hypothesize that these artifacts are an inherent side effect of perturbing the generation process via noise injection. 
To address this diversity-quality trade-off, in the following Sec.~\ref{subsec:method_scale-travel} we introduce a novel latent refinement method specifically designed for VAR.

\subsection{Scale-Travel: Latent Refinement for VAR}
\label{subsec:method_scale-travel}
In this section, we introduce a simple but effective refinement method designed for VAR. While condition-annealing (Sec.~\ref{subsec:method_condition_annealing}) significantly increases the diversity of VAR, it also introduces significant visual artifacts (second row of Fig.~\ref{fig:main_qualitative}). Inspired by diffusion model refinement techniques~\cite{meng2021sdedit,lugmayr2022repaint, bansal2023universal, yu2023freedom, ye2024tfg}, we introduce \textbf{Scale-Travel}, a playback strategy for repairing corrupted latents in VAR based on its \emph{multi-scale} nature. 
\begin{figure*}[t!]
    \centering
    \includegraphics[width=\textwidth]{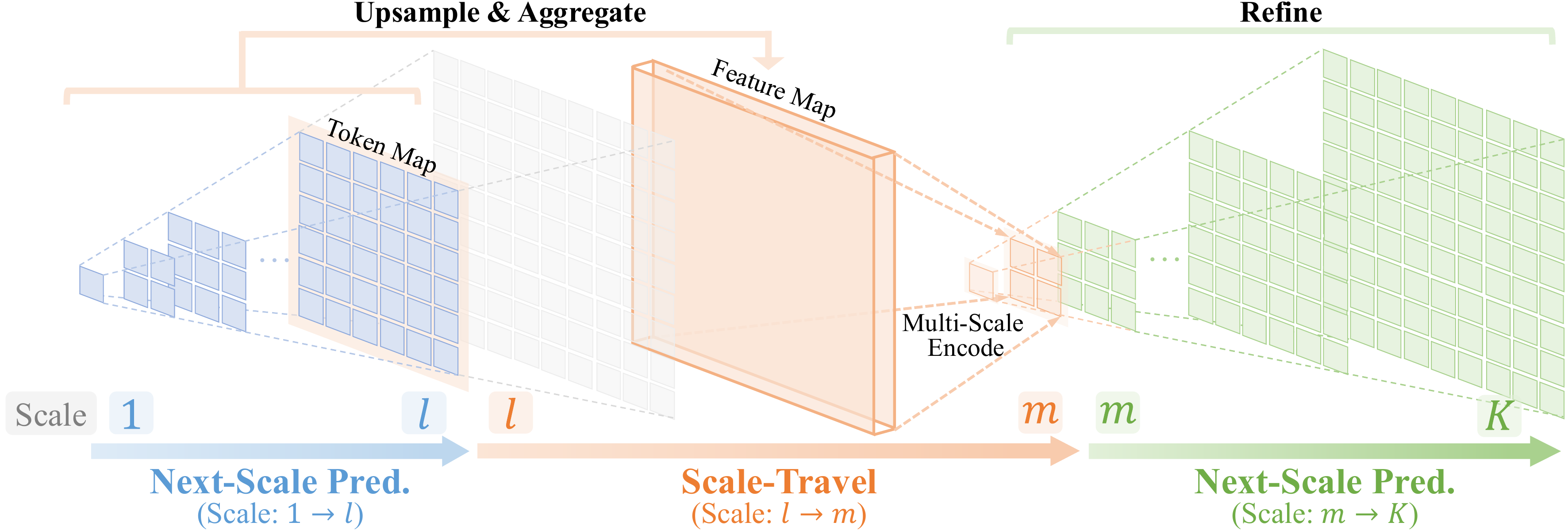}
    \vspace{-\baselineskip}
    \caption{\textbf{Latent Refinement via Scale-Travel.} We introduce scale-travel, a novel refinement strategy that leverages the multi-scale structure of VAR models. By reverting intermediate representations to a coarser scale and running generation without noise injection, clean and coherent details can be reconstructed. This process corrects visual artifacts and degradations while preserving the overall coarse structures.}
    \label{fig:scale_travel_pipeline}
    \vspace{-\baselineskip}
\end{figure*}

Recall from Eq.~\ref{eq:var-canvas} that in each stage $l$ the model forms a feature map $\mathbf{Z}_l$ by upsampling and summing the token grids $\{ \mathbf{r}_k \}_{k=1}^{l}$. After the noise injection of Sec.~\ref{subsec:method_condition_annealing}, $\mathbf{Z}_l$ may deviate from the model's image manifold. Scale-travel \emph{rewinds} the feature map $\mathbf{Z}_l$ to an earlier (\ie, coarser) scale $m < l$ and then resumes generation. To rewind the current feature map to an earlier scale, we utilize the \emph{multi-scale encoder} of VAR. Recall that during VAR training, an image $I$ is first encoded with a VQ-VAE~\cite{van2017vqvae} to obtain a dense feature map $\mathbf{Z}$, which is passed through a multi-scale encoder to be decomposed into scale-wise token grids

\begin{align}
    \mathbf{r}_{1:K} &= \texttt{MultiScaleEnc}(\mathbf{Z}, \mathbf{s}_{1:K})
\end{align}
where $\mathbf{s}_k$ is the grid size at stage $k$, as shown in Alg.~\ref{alg:multi-scale-encoding}. Here, we highlight a key property of VAR: early maps $\mathbf{Z}_l$ already contain the coarse layout and semantics of the final image (details included in the \textbf{supplementary}). While a similar observation has been presented in previous works~\cite{wang2025training, jiao2025flexvar}, we further employ this finding to apply multi-scale encoding to $\mathbf{Z}_l$---instead of ground-truth feature $\mathbf{Z}$---to reconstruct the lower-resolution grids $\Tilde{\mathbf{r}}_{1:m}$. We denote this technique as $\Tilde{\mathbf{r}}_{1:m} = \texttt{ScaleTravel}(\mathbf{Z}_l, \mathbf{s}_{1:m}, m)$ and compare the exact pseudocode to that of multi-scale encoding in Fig.~\ref{fig:alg_side_by_side}, showing that only a minor change in algorithm enables scale-travel. Once replayed to stage $m$, VAR's autoregressive generation proceeds from $k=m+1$ to $K$, effectively fixing the visual artifacts. We refer readers to the \textbf{supplementary} for more details and visualization on the trajectory of scale-travel. The full latent refinement technique is shown in Fig.~\ref{fig:scale_travel_pipeline} and can be written as
{\small
\begin{align}
    & \text{\texttt{Scale-Travel:}} \quad \Tilde{\mathbf{r}}_{1:m} = \texttt{ScaleTravel}(\mathbf{Z}_l, \mathbf{s}_{1:m}, m) \\
    & \text{\texttt{Refine:}} \quad \hat{\mathbf{r}}_{m+1:K} \sim \prod_{k=m+1}^K p(\mathbf{r}_k | \Tilde{\mathbf{r}}_1,\dots, \Tilde{\mathbf{r}}_m, \mathbf{r}_{m+1},\dots,\mathbf{r}_{k-1})
\end{align}}

\begin{figure*}[h]
  \centering
  \footnotesize
  \begin{minipage}[t]{0.48\textwidth}
    \centering
    \begin{algorithm}[H]
      \caption{Multi-Scale Encode}
      \label{alg:multi-scale-encoding}
      \KwParam{Latent $\mathbf{Z}$, Scales $\mathbf{s}_{1:K}$}
      \vspace{1.8em}
      \myalg{\colorbox{yellow!10}{\algo{$\mathcolor{blue}{\mathbf{Z}}$, $\mathbf{s}_{1:K}$}}}{
        $\mathbf{R}\leftarrow[]$\;
        \For{$k=1,2,\dots,\mathcolor{blue}{K}$}{
          $\mathbf{r}_k \leftarrow \mathcal{Q}(\texttt{down}(\mathcolor{blue}{\mathbf{Z}} - \mathbf{Z}_{k-1}, \mathbf{s}_k))$\;
          $\mathbf{R}.\text{append}(\mathbf{r}_k)$\;
          $\mathbf{Z}_k \leftarrow \sum_{i=1}^{k}\texttt{up}(\mathbf{r}_i, \mathbf{s}_K )$\;
        }
        \Return $\mathbf{R}= \mathbf{r}_{1:\mathcolor{blue}{K}}$
      }
    \end{algorithm}
  \end{minipage}
  \hfill
  \begin{minipage}[t]{0.5\textwidth}
    \centering
    \begin{algorithm}[H]
      \caption{Scale-Travel}
      \label{alg:scale-travel}
      \KwParam{Latent $\mathbf{Z}_l$, Scales $\mathbf{s}_{1:m}$, Target $m$}
      \vspace{0.6em}
      \myalg{\colorbox{orange!10}{\algost{$\mathcolor{red}{\mathbf{Z}_l}$, $\mathbf{s}_{1:m}$, $\mathcolor{red}{m}$}}}{
        $\mathbf{R}\leftarrow[]$\;
        \For{$k=1,2,\dots,\mathcolor{red}{m}$}{
          $\mathbf{r}_k \leftarrow \mathcal{Q}(\texttt{down}(\mathcolor{red}{\mathbf{Z}_l} - \mathbf{Z}_{k-1}, \mathbf{s}_k))$\;
          $\mathbf{R}.\text{append}(\mathbf{r}_k)$\;
          $\mathbf{Z}_k \leftarrow \sum_{i=1}^{k}\texttt{up}(\mathbf{r}_i, \mathbf{s}_K )$\;
        }
        \Return $\mathbf{R}= \mathbf{r}_{1:\mathcolor{red}{m}}$
      }
    \end{algorithm}
  \end{minipage}

  \caption{\textbf{Scale-Travel Pseudocode.} Side-by-side pseudocode for \textbf{(left)} multi-scale encoding process in VAR~\cite{tian2024visual} and \textbf{(right)} our proposed scale-travel step in Sec.~\ref{subsec:method_scale-travel}. $\texttt{up}(\cdot,\cdot)$ and $\texttt{down}(\cdot,\cdot)$ denote bilinear upsampling and downsampling, respectively.}
  \label{fig:alg_side_by_side}
\end{figure*}

\begin{figure}[t!]
    \centering
    \includegraphics[width=\linewidth]{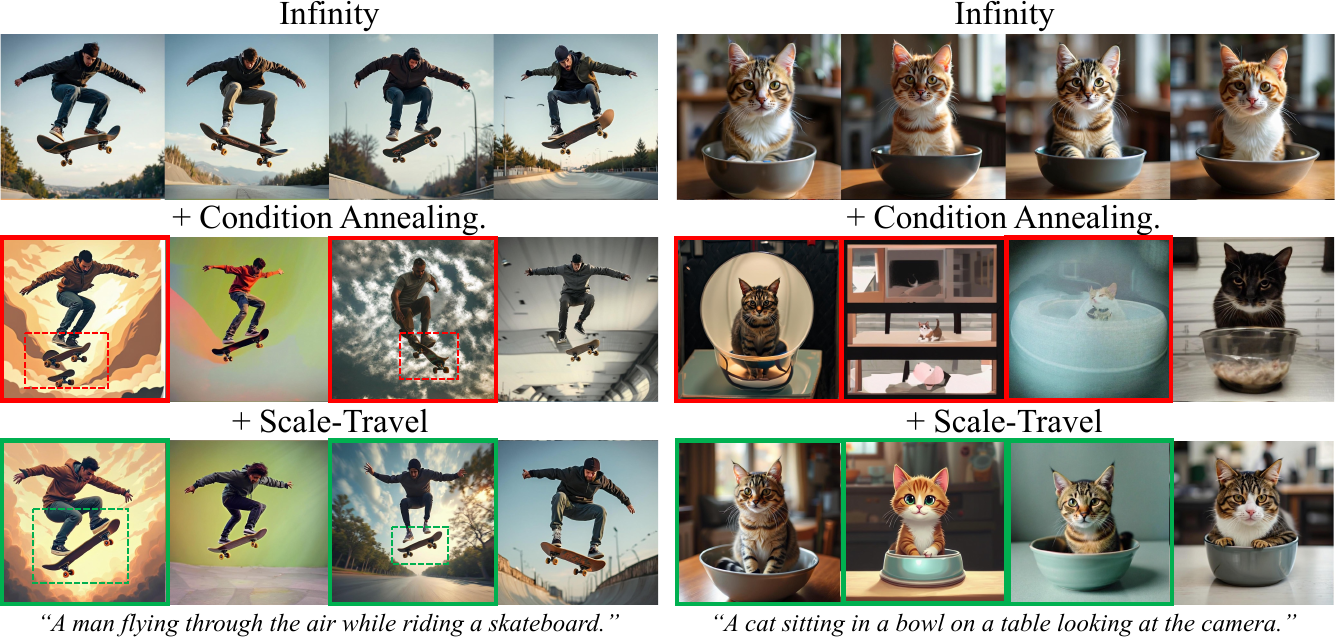}
    \vspace{-1.5\baselineskip}
    \caption{\textbf{Qualitative Comparisons on Image Generation.} Infinity produces images with little variation and diversity (row 1). Injecting noise into the text-embedding increases diversity (row 2) but results in visual artifacts (\textcolor{red}{red box}). Applying our \textbf{scale-travel} refinement technique (row 3) fixes these visual artifacts (\textcolor{darkgreen}{green box}) while retaining diversity.}
    \label{fig:main_qualitative}
    \vspace{-\baselineskip}
\end{figure}

\begin{figure*}[!h]
    \centering
   \begin{minipage}{\textwidth}
     \centering
     \includegraphics[width=\linewidth]{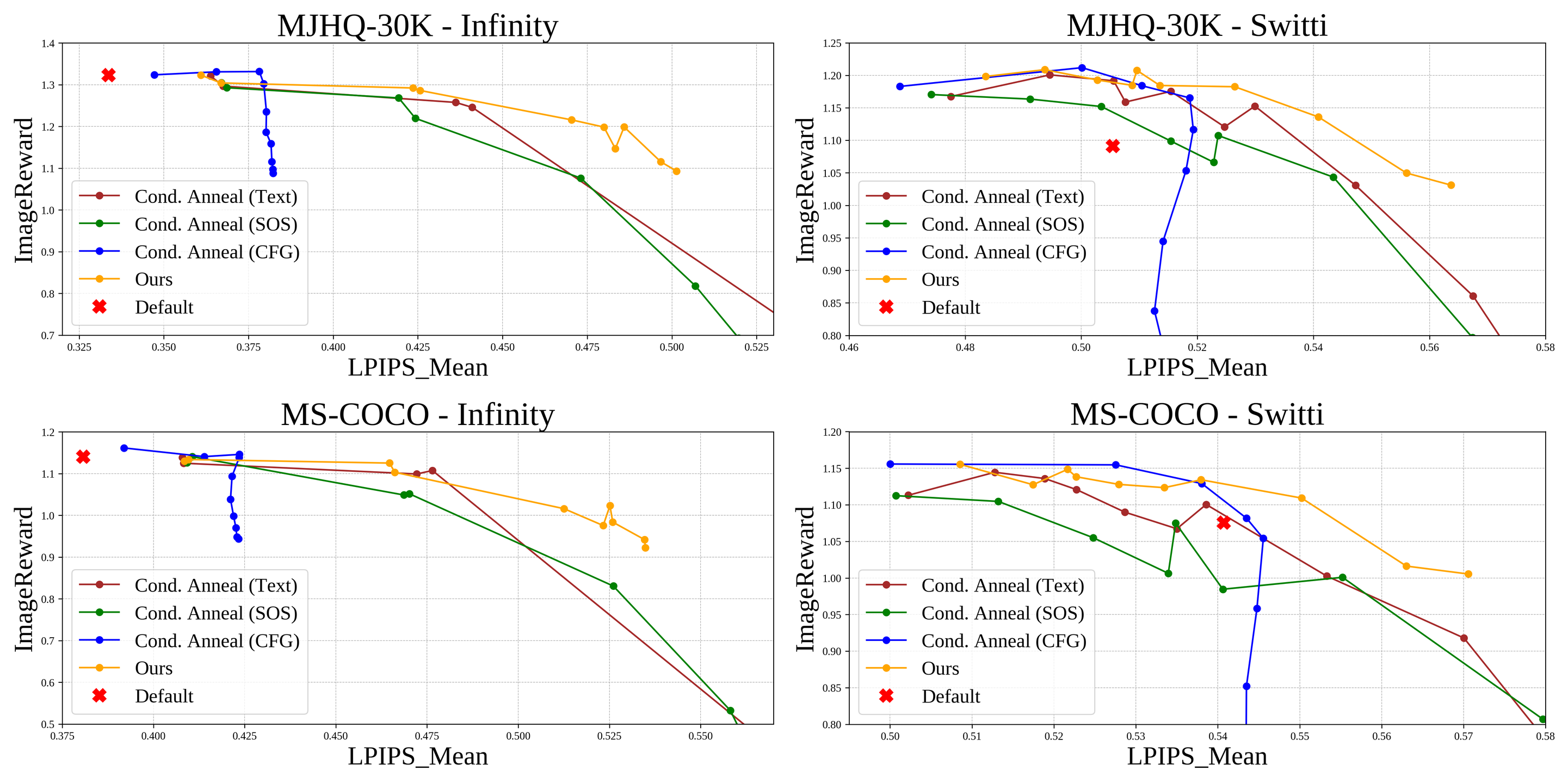}
   \end{minipage}
 \caption{\textbf{Pareto Fronts for Diversity-Quality Trade-Off on Infinity~\cite{han2024infinity} and Switti~\cite{voronov2024switti}}. Each curve is obtained by varying hyperparameters for each method: noise scales for SOS and CADS, guidance schedules for CFG, and target scales for \methodname{} (Ours). The default setting of each model is included as a reference. 
 On both MJHQ-30K~\cite{li2024mjhq} and MS-COCO~\cite{lin2014mscoco}, \methodname{} consistently sets the Pareto frontier, demonstrating the best balance between diversity and quality in VAR.}
 \label{fig:pareto}
\end{figure*}

\section{Results}
\label{sec:results}
\vspace{-0.5\baselineskip}
In this section, we present experimental results of \methodname{}, comparing diversity and quality with existing VAR diversity enhancement techniques.

\vspace{-0.5\baselineskip}
\subsection{Evaluation Setup}
\label{subsec:experimental-setup}
\paragraph{VAR Models.} 
For text-conditioned image generation VAR models, we use Infinity~\cite{han2024infinity} and Switti~\cite{voronov2024switti}. Infinity trains a bitwise transformer for scaling the vocabulary to infinite size, and Switti improves the efficiency and scalability of VAR~\cite{tian2024visual} by addressing training instability and using non-causal self-attention for faster sampling. 
For text-conditioned video generation with VAR models, we use InfinityStar-8B~\cite{liu2025:infinitystar}, which generates 480p videos with 81 frames. Since InfinityStar follows a two-stage pipeline, first generating keyframes and then performing temporal generation, we apply \methodname{} to the keyframe generation stage.

\vspace{-0.5\baselineskip}
\paragraph{Benchmarks and Evaluation Metrics.} 
For text-conditioned image generation, we evaluate on MS-COCO~\cite{lin2014mscoco} and MJHQ-30K~\cite{li2024mjhq} datasets, each of which consists of image-caption pairs. Note that MS-COCO consists of realistic photos, while MJHQ-30K is constructed with images generated by Midjourney~\cite{Midjourney2022}. 
For text-conditioned video generation, we evaluate all methods on VBench~\cite{huang2024vbench,zheng2025:vbench2} which evaluates video generative models across multiple dimensions. We generate videos based on VBench-2.0~\cite{zheng2025:vbench2} prompts. 
We use multiple metrics to measure image/video quality and diversity as follows:

\begin{itemize}[leftmargin=8pt]
    \item \underline{Image Diversity}: \textbf{LPIPS-MPD}~\cite{ahn2024pag} computes the mean pairwise LPIPS~\cite{zhang2018lpips} distance between images to measure per-prompt diversity.
    \item \underline{Image Diversity}: \textbf{Vendi-Score}~\cite{friedman2022vendi} computes the effective number of ``dissimilar'' elements in a sample of images. Following CADS~\cite{sadat2023cads}, we use SSCD~\cite{pizzi2022self} to measure pairwise dissimilarities.
    \item \underline{Image Quality}: \textbf{ImageReward}~\cite{xu2023imagereward} evaluates image quality and alignment based on human preference.
    \item \underline{Image Quality \& Diversity}: \textbf{FID}~\cite{heusel2017fid} simultaneously captures both diversity and quality with respect to a reference set of images.
    \item 
    \underline{Video Diversity}: \textbf{VBench-2.0 Diversity}~\cite{zheng2025:vbench2} computes video diversity using inter-sample style and content variation with a pretrained VGG-19~\cite{simonyan2015vgg}.
    \item 
    \underline{Video Quality}: \textbf{VBench Quality Score}~\cite{huang2024vbench} is used to evaluate video generation quality based on frame-wise metrics and temporal metrics. Results for each VBench category are provided in the \textbf{supplementary} material.
    
\end{itemize}

\subsection{Pareto Fronts on Diversity-Quality Trade-Off}
\label{subsec:pareto_diversity_quality}
We assess \methodname’s diversity–quality balance for Infinity~\cite{han2024infinity} and Switti~\cite{voronov2024switti} by plotting Pareto fronts on the metrics above.

\vspace{-0.25\baselineskip}
\paragraph{Settings.}
We compare \methodname{} against the three diversity enhancement techniques from Sec.~\ref{subsec:method_condition_annealing}: CFG scheduling and condition-annealing (\ie, noise injection) with text embeddings and \sos~tokens. For each dataset~\cite{li2024mjhq, lin2014mscoco}, we sample 100 prompts and generate 10 images per prompt to trace each Pareto curve in Fig.~\ref{fig:pareto}. We use 1000 prompts and generate 10 images per prompt for Tab.~\ref{tab:main_quantitative}. Our \methodname{} pairs scale-travel with text-embedding annealing, which we observed to be the most effective option.
Specifically, we fix $k_{\text{min}}=1$ and vary $k_{\text{max}}\in\{2, 3, 4, 5, 6\}$ with the maximum noise scales $\alpha(1), \beta(1)\in\{0.5, 1.0\}$ for condition-annealing on both text embedding and \sos. For CFG scheduling, we set $k_{\text{max}} \in \{1, 2, \dots, 10\}$ and use each model’s default guidance scale as $w_K$. For \methodname{}, scale-travel starts at scale $l=8$, where we observed VAR forms an image's global structure and semantics. We use a constant scheduler for CFG scheduling and a cosine scheduler for condition-annealing. More details and ablation studies on the scheduler, including specific hyperparameters used, are provided in the \textbf{supplementary}. 

\vspace{-0.5\baselineskip}
\paragraph{Results.}
Fig.~\ref{fig:pareto} shows the Pareto fronts for all methods. Across both benchmarks and different metrics, \methodname{} consistently defines the frontier, underscoring that our scale-travel effectively refines the output image quality while retaining the diversity introduced by condition-annealing.
CFG scheduling increases the diversity but also significantly degrades the image quality. 
Injecting noise into the text embedding or \sos~token enhances diversity beyond what CFG scheduling can offer, by preventing the dominance of the conditioning signal during early generation steps. Among these condition-annealing techniques, noise injection to the text embedding yields the best diversity-quality trade-off, yet results in clear quality degradation---as shown in the sharp drop in the Pareto front as diversity increases. Our \methodname{} overcomes this drawback by applying scale-travel to recover image quality while preserving the added diversity, thereby achieving the Pareto frontier.

\begin{table*}[!th]
\setlength{\tabcolsep}{5pt}
\scriptsize
\centering
\begin{tabularx}{\textwidth}{@{}>{\raggedright\arraybackslash}X *{8}{c}@{}}
    \toprule
    \multirow{2}{*}{\textbf{Method}} & \multicolumn{4}{c}{\textbf{MJHQ-30K~\cite{li2024mjhq}}} & \multicolumn{4}{c}{\textbf{MS-COCO~\cite{lin2014mscoco}}} \\
    \cmidrule(lr){2-5} \cmidrule(lr){6-9}
    & \textbf{FID$\downarrow$} & \textbf{IR$\uparrow$} & \textbf{MPD$_{\text{L}}$ $\uparrow$} & \textbf{Vendi $\uparrow$}
     & \textbf{FID$\downarrow$} & \textbf{IR$\uparrow$}& \textbf{MPD$_{\text{L}}$ $\uparrow$} & \textbf{Vendi $\uparrow$} \\
    \midrule
    Infinity~\cite{han2024infinity} & 19.16  & 1.22& 0.33 & 4.10 & 37.37  & 1.16& 0.37 & 4.41 \\
    \quad + Cond. Anneal. & 15.53  & 0.49 & 0.56 & 7.68 & 22.51  & 0.39 & 0.58 & 8.13 \\
    \rowcolor{gray!12}\quad + Scale-Travel & 15.28  & 1.08& 0.48 & 6.07 & 28.97 & 1.04 & 0.52 & 6.68 \\
    \midrule
    Switti~\cite{voronov2024switti}& 16.18  & 1.11 & 0.44 & 4.79 & 25.92 & 1.18 & 0.46 & 5.32 \\
    \quad + Cond. Anneal. & 21.34  & 0.63& 0.58 & 7.34 & 25.26 & 0.78 & 0.59 & 7.74 \\
    \rowcolor{gray!12}\quad + Scale-Travel & 17.30 & 0.93 & 0.56 & 6.68 & 23.47 & 1.05 & 0.57 & 7.22 \\
    \bottomrule
    
\end{tabularx}
\vspace{0.5\baselineskip}
\caption{\textbf{Quantitative Comparisons on Image Generation.} We evaluate four metrics: ImageReward (IR)~\cite{xu2023imagereward} for quality, FID~\cite{heusel2017fid} for both quality and diversity, and LPIPS-MPD (MPD$_{\text{L}}$)~\cite{zhang2018lpips}, Vendi~\cite{friedman2022vendi} for diversity. 
For both Switti~\cite{voronov2024switti} and Infinity~\cite{han2024infinity}, condition annealing enhances diversity but noticeably degrades quality. In contrast, \methodname's scale-travel refinement finds a more favorable trade-off, balancing quality and diversity.}
\vspace{-2\baselineskip}
\label{tab:main_quantitative}
\end{table*}

\subsection{Quantitative and Qualitative Comparisons}
\label{subsec:main-results}
We additionally report quantitative and qualitative evaluations of \methodname{}. For this, we use the optimal hyperparameter settings for each method, selected based on a toy experiment, which we report in detail in the \textbf{supplementary}. Since we find that injecting noise at every scale results in severe quality degradation, we adopt a scheduling strategy with a threshold $k_{\text{max}} < K$~\cite{sadat2023cads}.

\paragraph{Image Generation.} As shown in Tab.~\ref{tab:main_quantitative}, our method generalizes to different text-to-image VARs. Starting from the original model, condition annealing significantly enhances diversity. Notably, adding noise to the text embedding (Cond. Anneal.) yields a substantial diversity gain in LPIPS-MPD~\cite{ahn2024pag} on the MJHQ-30K~\cite{li2024mjhq} dataset, $+69.7\%$ improvement over the baseline Infinity. However, this comes with a clear trade-off in terms of image quality. Our scale-travel method significantly enhances image quality from noisy samples, achieving a $+120.41\%$ improvement on ImageReward~\cite{xu2023imagereward}, while incurring only a modest reduction in diversity ($-14.29\%$ in LPIPS-MPD). Compared to the base model, \methodname{} yields a $+45.45\%$ improvement in LPIPS-MPD with minimal quality degradation for ImageReward ($-11.48\%$). A similar trend is also observed for Switti. This underscores the effectiveness of our refinement technique: starting from diverse but low-quality samples, scale-travel improves image quality while retaining a substantial diversity gain over standard text-conditioned VARs, with only a minor decrease in quality.

Moreover, in Fig.~\ref{fig:main_qualitative}, we present qualitative results from each stage of our \methodname{} pipeline. The outputs from the default model (row 1) exhibit a severe lack of diversity across the same text condition. After injecting noise into the text conditions, we observe a substantial increase in diversity compared to the previous stage (row 2). However, this also leads to noticeable quality degradation, often failing to generate clear objects or preserve fine details.
The last row shows the final output of \methodname{}, demonstrating that our method not only corrects the corrupted details introduced by condition-annealing but also preserves the high-level semantics. More qualitative results are included in the \textbf{supplementary}.

\vspace{-0.75\baselineskip}

\begin{figure}[t!]
\centering
\setlength{\tabcolsep}{0.6pt}
\renewcommand{\arraystretch}{0.35}

\begin{minipage}{0.495\textwidth}
\centering

\newcommand{\clipid}{0}
\newcommand{\vidframe}[3]{figures/results/video_samples/#1/\clipid/#2_#3.jpg}
\newcommand{\vidprompt}{figures/results/video_samples/prompts/\clipid.txt}
\begin{tabular}{>{\centering\arraybackslash}p{0.9em} cccc >{\centering\arraybackslash}p{0.5em}}

{} & \scriptsize Frame 1 & \scriptsize Frame 26 & \scriptsize Frame 51 & \scriptsize Frame 75 & {} \\

\multirow{3}{*}[ 1.5ex ]{\rotatebox{90}{\scriptsize InfinityStar}} &
\includegraphics[width=0.22\linewidth]{\vidframe{base}{0}{0}} &
\includegraphics[width=0.22\linewidth]{\vidframe{base}{0}{1}} &
\includegraphics[width=0.22\linewidth]{\vidframe{base}{0}{2}} &
\includegraphics[width=0.22\linewidth]{\vidframe{base}{0}{3}} &
\raisebox{5.2ex}{\rotatebox{270}{\tiny Seed~1}} \\
&
\includegraphics[width=0.22\linewidth]{\vidframe{base}{1}{0}} &
\includegraphics[width=0.22\linewidth]{\vidframe{base}{1}{1}} &
\includegraphics[width=0.22\linewidth]{\vidframe{base}{1}{2}} &
\includegraphics[width=0.22\linewidth]{\vidframe{base}{1}{3}} &
\raisebox{5.2ex}{\rotatebox{270}{\tiny Seed~2}} \\
&
\includegraphics[width=0.22\linewidth]{\vidframe{base}{2}{0}} &
\includegraphics[width=0.22\linewidth]{\vidframe{base}{2}{1}} &
\includegraphics[width=0.22\linewidth]{\vidframe{base}{2}{2}} &
\includegraphics[width=0.22\linewidth]{\vidframe{base}{2}{3}} &
\raisebox{5.2ex}{\rotatebox{270}{\tiny Seed~3}} \\

\multirow{3}{*}[ 3.0ex ]{\rotatebox{90}{\textbf{\scriptsize \methodname{}}}} &
\includegraphics[width=0.22\linewidth]{\vidframe{ours}{0}{0}} &
\includegraphics[width=0.22\linewidth]{\vidframe{ours}{0}{1}} &
\includegraphics[width=0.22\linewidth]{\vidframe{ours}{0}{2}} &
\includegraphics[width=0.22\linewidth]{\vidframe{ours}{0}{3}} &
\raisebox{5.2ex}{\rotatebox{270}{\tiny Seed~1}} \\
&
\includegraphics[width=0.22\linewidth]{\vidframe{ours}{1}{0}} &
\includegraphics[width=0.22\linewidth]{\vidframe{ours}{1}{1}} &
\includegraphics[width=0.22\linewidth]{\vidframe{ours}{1}{2}} &
\includegraphics[width=0.22\linewidth]{\vidframe{ours}{1}{3}} &
\raisebox{5.2ex}{\rotatebox{270}{\tiny Seed~2}} \\
&
\includegraphics[width=0.22\linewidth]{\vidframe{ours}{2}{0}} &
\includegraphics[width=0.22\linewidth]{\vidframe{ours}{2}{1}} &
\includegraphics[width=0.22\linewidth]{\vidframe{ours}{2}{2}} &
\includegraphics[width=0.22\linewidth]{\vidframe{ours}{2}{3}} &
\raisebox{5.2ex}{\rotatebox{270}{\tiny Seed~3}} \\

\end{tabular}

\vspace{2pt}
\parbox{0.98\linewidth}{\centering\scriptsize ``\emph{\input{\vidprompt}}''$^\dagger$}

\end{minipage}
\hfill
\begin{minipage}{0.495\textwidth}
\centering
\newcommand{\clipid}{1}
\newcommand{\vidframe}[3]{figures/results/video_samples/#1/\clipid/#2_#3.jpg}
\newcommand{\vidprompt}{figures/results/video_samples/prompts/\clipid.txt}
\begin{tabular}{>{\centering\arraybackslash}p{0.9em} cccc >{\centering\arraybackslash}p{0.5em}}

{} & \scriptsize Frame 1 & \scriptsize Frame 26 & \scriptsize Frame 51 & \scriptsize Frame 75 & {} \\

\multirow{3}{*}[ 1.5ex ]{\rotatebox{90}{\scriptsize InfinityStar}} &
\includegraphics[width=0.22\linewidth]{\vidframe{base}{0}{0}} &
\includegraphics[width=0.22\linewidth]{\vidframe{base}{0}{1}} &
\includegraphics[width=0.22\linewidth]{\vidframe{base}{0}{2}} &
\includegraphics[width=0.22\linewidth]{\vidframe{base}{0}{3}} &
\raisebox{5.2ex}{\rotatebox{270}{\tiny Seed~1}} \\
&
\includegraphics[width=0.22\linewidth]{\vidframe{base}{1}{0}} &
\includegraphics[width=0.22\linewidth]{\vidframe{base}{1}{1}} &
\includegraphics[width=0.22\linewidth]{\vidframe{base}{1}{2}} &
\includegraphics[width=0.22\linewidth]{\vidframe{base}{1}{3}} &
\raisebox{5.2ex}{\rotatebox{270}{\tiny Seed~2}} \\
&
\includegraphics[width=0.22\linewidth]{\vidframe{base}{2}{0}} &
\includegraphics[width=0.22\linewidth]{\vidframe{base}{2}{1}} &
\includegraphics[width=0.22\linewidth]{\vidframe{base}{2}{2}} &
\includegraphics[width=0.22\linewidth]{\vidframe{base}{2}{3}} &
\raisebox{5.2ex}{\rotatebox{270}{\tiny Seed~3}} \\

\multirow{3}{*}[ 3.0ex ]{\rotatebox{90}{\textbf{\scriptsize \methodname{}}}} &
\includegraphics[width=0.22\linewidth]{\vidframe{ours}{0}{0}} &
\includegraphics[width=0.22\linewidth]{\vidframe{ours}{0}{1}} &
\includegraphics[width=0.22\linewidth]{\vidframe{ours}{0}{2}} &
\includegraphics[width=0.22\linewidth]{\vidframe{ours}{0}{3}} &
\raisebox{5.2ex}{\rotatebox{270}{\tiny Seed~1}} \\
&
\includegraphics[width=0.22\linewidth]{\vidframe{ours}{1}{0}} &
\includegraphics[width=0.22\linewidth]{\vidframe{ours}{1}{1}} &
\includegraphics[width=0.22\linewidth]{\vidframe{ours}{1}{2}} &
\includegraphics[width=0.22\linewidth]{\vidframe{ours}{1}{3}} &
\raisebox{5.2ex}{\rotatebox{270}{\tiny Seed~2}} \\
&
\includegraphics[width=0.22\linewidth]{\vidframe{ours}{2}{0}} &
\includegraphics[width=0.22\linewidth]{\vidframe{ours}{2}{1}} &
\includegraphics[width=0.22\linewidth]{\vidframe{ours}{2}{2}} &
\includegraphics[width=0.22\linewidth]{\vidframe{ours}{2}{3}} &
\raisebox{5.2ex}{\rotatebox{270}{\tiny Seed~3}} \\

\end{tabular}

\vspace{2pt}
\parbox{0.98\linewidth}{\centering\scriptsize ``\emph{\input{\vidprompt}}''$^\dagger$}

\end{minipage}

\vspace{-0.7em}
\caption{\textbf{Qualitative Comparisons on Video Generation.} InfinityStar~\cite{liu2025:infinitystar} produces videos with little variation and diversity (row 1-3). 
Injecting noise into the text-embedding and applying our \textbf{scale-travel} refinement technique (row 4-6) can significantly increase per-prompt video diversity. $^\dagger$ denotes using augmented prompt for generation.}
\vspace{-\baselineskip}
\label{fig:video_qualitative}
\end{figure}

\begin{wraptable}[6]{r}{0.5\textwidth}
\vspace{-2.0\baselineskip}
\scriptsize
\centering
\begin{tabularx}{\linewidth}{@{}>{\raggedright\arraybackslash}X c c@{}}
    \toprule
    \textbf{Method} & \textbf{Diversity} & \textbf{Quality Score} \\
    \midrule
    InfinityStar~\cite{liu2025:infinitystar} & 0.270 & 0.806 \\
    \, + Cond. Anneal. & 0.627 & 0.794 \\
    \rowcolor{gray!12}\, + Scale-Travel & 0.487 & 0.804 \\
    \bottomrule
\end{tabularx}
\vspace{-\baselineskip}
\caption{\textbf{Quantitative results on InfinityStar~\cite{liu2025:infinitystar} using VBench~\cite{huang2024vbench,zheng2025:vbench2}.}}
\label{tab:video-quantitative}
\end{wraptable}
\paragraph{Video Generation.} 
As shown in Tab.~\ref{tab:video-quantitative}, \methodname{} improves diversity while largely preserving overall VBench performance, yielding a more favorable trade-off than condition annealing. Compared to the baseline InfinityStar, condition annealing delivers a larger diversity increase ($+132.2\%$) but reduces the quality score. In contrast, \methodname{} raises diversity from \(0.270\) to \(0.487\) (\(+80.4\%\)) with a negligible change in quality ($-0.002$ points).

In addition, Fig.~\ref{fig:video_qualitative} provides qualitative comparisons between the original InfinityStar~\cite{liu2025:infinitystar} and our \methodname{} pipeline across multiple random seeds. For a fixed prompt, the original model (rows~1--3) exhibits limited diversity, producing nearly identical videos across different seeds. In contrast, \methodname{} (rows~4--6) yields substantially more varied outputs over the entire sequence. The differences are particularly evident in color attributes and scene composition: InfinityStar~\cite{liu2025:infinitystar} tends to reproduce the same background and object layout with minimal variation, while our method generates diverse videos and maintains overall coherence. 
More qualitative results are included in the \textbf{supplementary}.

\subsection{Comparison with Alternative Diversity Controls}
\label{subsec:other_diversity_controls}
\paragraph{CFG Scale, Temperature, and Nucleus Sampling.} We also provide additional results under varying CFG weights, temperatures, and nucleus sampling parameters~\cite{holtzman2020:nucleus}. As shown in Tab.~\ref{tab:extra_result_cfg_ablation}, \methodname{} consistently improves the diversity-quality tradeoff across all CFG weights. As shown in Tab.~\ref{tab:extra_result_temperature} and Tab.~\ref{tab:extra_result_top_p}, \methodname{} also yields a better diversity-quality tradeoff than simply varying the temperature (Tab.~\ref{tab:extra_result_temperature}) or using nucleus sampling (Tab.~\ref{tab:extra_result_top_p}). 

\begin{table*}[!t]
\scriptsize
\centering

\begin{tabularx}{\linewidth}{>{\centering\arraybackslash}m{0.03\textwidth}>{\centering\arraybackslash}m{0.15\textwidth}YYY @{\hspace{2pt}}||@{\hspace{2pt}} 
>{\centering\arraybackslash}m{0.03\textwidth}>{\centering\arraybackslash}m{0.15\textwidth}YYY}

\toprule
\textbf{$\omega$} & \textbf{Method} & \textbf{MPD $\uparrow$} & \textbf{Vendi $\uparrow$} & \textbf{IR $\uparrow$} & \textbf{$\omega$} & \textbf{Method} & \textbf{MPD $\uparrow$} & \textbf{Vendi $\uparrow$} & \textbf{IR $\uparrow$} \\
\midrule

\multirow{2}{*}{$2.0^\dagger$} & $\text{Infinity}$ & 0.334 & 4.031 & 1.323 & \multirow{2}{*}{2.0} & Switti & 0.465 & 5.262 & 1.100 \\
& \methodname{} & 0.486 & 6.087 & 1.199 & & \methodname{} & 0.568 & 7.034 & 0.919 \\
\midrule
\multirow{2}{*}{4.0} & Infinity & 0.324 & 3.862 & 1.367 & \multirow{2}{*}{4.0} & Switti & 0.448 & 4.921 & 1.165 \\
& \methodname{} & 0.486 & 6.127 & 1.215 & & \methodname{} & 0.565 & 6.705 & 1.035 \\
\midrule
\multirow{2}{*}{6.5} & Infinity & 0.320 & 3.804 & 1.384 & \multirow{2}{*}{$6.0^\dagger$} & $\text{Switti}$ & 0.444 & 4.744 & 1.210 \\
& \methodname{} & 0.481 & 6.101 & 1.234 & & \methodname{} & 0.564 & 6.609 & 1.031 \\
\midrule
\multirow{2}{*}{9.0} & Infinity & 0.320 & 3.770 & 1.406 & \multirow{2}{*}{9.0} & Switti & 0.438 & 4.668 & 1.225 \\
& \methodname{} & 0.476 & 6.019 & 1.239 & & \methodname{} & 0.562 & 6.557 & 0.974 \\
\bottomrule
\end{tabularx}
\vspace{0.5\baselineskip}
\caption{\textbf{\methodname{} with Varying CFG Weights}. We vary the CFG guidance weights $\omega$ for the VAR models and \methodname{}. \methodname{} consistently improves the diversity-quality tradeoff across all CFG weights.
}
\label{tab:extra_result_cfg_ablation}
\vspace{-1.5\baselineskip}
\end{table*}
\begin{table*}[!t]
\scriptsize
\centering

\begin{tabularx}{\linewidth}{>{\qquad\arraybackslash}m{0.2\textwidth}YYY ||  
>{\qquad\arraybackslash}m{0.2\textwidth}YYY}
\toprule
\textbf{Infinity}~\cite{han2024infinity} & \textbf{MPD $\uparrow$} & \textbf{Vendi $\uparrow$} & \textbf{IR $\uparrow$} & \textbf{Switti}~\cite{voronov2024switti} & \textbf{MPD $\uparrow$} & \textbf{Vendi $\uparrow$} & \textbf{IR $\uparrow$} \\
\midrule
$\tau=0.5$ & 0.284 & 3.500 & 1.305 & $\tau=0.5$ & 0.437 & 4.483 & 1.204 \\
$\tau=1^\dagger$ & 0.334 & 4.031 & 1.323 & $\tau=1^\dagger$ & 0.444 & 4.744 & 1.210 \\
$\tau=2$ & 0.375 & 4.577 & 1.331 & $\tau=2$ & 0.453 & 5.014 & 1.095 \\
$\tau=5$ & 0.414 & 4.813 & 1.108 & $\tau=5$ & 0.477 & 5.613 & 0.813 \\
$\tau=10$ & 0.459 & 4.954 & -0.430 & $\tau=10$ & 0.487 & 5.897 & 0.619 \\
\midrule
\rowcolor{gray!12}\methodname{} & 0.470 & 5.833 & 1.216 & \methodname{} & 0.563 & 6.609 & 1.031 \\
\bottomrule
\end{tabularx}
\vspace{0.5\baselineskip}
\caption{\textbf{VAR models with varying temperatures.} While increasing $\tau$ leads to larger diversity, the image quality (IR) is significantly degraded. \methodname{} achieves the highest diversity while at the same time maintaining image quality. 
}
\label{tab:extra_result_temperature}
\vspace{-1.5\baselineskip}
\end{table*}
\begin{table*}[!t]
\scriptsize
\centering
\begin{tabularx}{\linewidth}{>{\qquad\arraybackslash}m{0.2\textwidth}YYY||>{\qquad\arraybackslash}m{0.2\textwidth}YYY}
\toprule

\textbf{Infinity}~\cite{han2024infinity} & \textbf{MPD $\uparrow$} & \textbf{Vendi $\uparrow$} & \textbf{IR $\uparrow$} & \textbf{Switti}~\cite{voronov2024switti} & \textbf{MPD $\uparrow$} & \textbf{Vendi $\uparrow$} & \textbf{IR $\uparrow$} \\
\midrule
$p=0.9$ & 0.323 & 3.934 & 1.320 & $p=0.9$ & 0.441 & 4.714 & 1.206 \\
$p=0.95$ & 0.331 & 4.016 & 1.329 & $p=0.95^\dagger$ & 0.444 & 4.744 & 1.210 \\
$p=0.97^\dagger$ & 0.334 & 4.031 & 1.323 & $p=0.97$ & 0.444 & 4.763 & 1.205 \\
$p=1.0$ & 0.335 & 4.071 & 1.330 & $p=1.0$ & 0.445 & 4.771 & 1.200 \\
\midrule
\rowcolor{gray!12}\methodname{} & 0.470 & 5.833 & 1.216 & \methodname{} & 0.563 & 6.609 & 1.031 \\
\bottomrule
\end{tabularx}
\vspace{0.5\baselineskip}
\caption{\textbf{VAR models with varying nucleus sampling parameters~\cite{holtzman2020:nucleus}.} Varying $p$ does not significantly impact the diversity. \methodname{} achieves significantly higher diversity without significantly degrading the image quality.}
\label{tab:extra_result_top_p}
\vspace{-1.5\baselineskip}
\end{table*}
\begin{figure}[!t]
    \centering
    \setlength{\tabcolsep}{1pt}
    \renewcommand{\arraystretch}{1.0}
    \newcolumntype{Y}{>{\centering\arraybackslash}X}
    \def\promptidx{0}
    \def\imgA{0}
    \def\imgB{1}
    \def\imgC{2}
    \def\imgD{3}
    \def\imgE{4}
    \def\imgF{5}

    \newcommand{\imgpath}[2]{figures/results/prompt_rewriting/#1/#2}

    \newcommand{\methodrow}[3]{
        \includegraphics[width=\linewidth]{\imgpath{\promptidx}{#1/#2.jpg}} &
        \includegraphics[width=\linewidth]{\imgpath{\promptidx}{#1/#3.jpg}}
    }
    
    \begin{tabularx}{\linewidth}{@{}*{2}{Y}|*{2}{Y}|*{2}{Y}|*{2}{Y}@{}}
        \toprule
        \multicolumn{2}{c|}{Infinity} &
        \multicolumn{2}{c|}{Infinity+P.R.} &
        \multicolumn{2}{c|}{\methodname{}} &
        \multicolumn{2}{c}{\methodname{}+P.R.} \\
        \midrule
        \multicolumn{8}{p{\linewidth}}{\centering\scriptsize Original prompt: ``\texttt{\input{figures/results/prompt_rewriting/\promptidx/prompt.txt}}\unskip''} \\
        \midrule
        \methodrow{baseline}{\imgA}{\imgB} &
        \methodrow{prompt_rewrite}{\imgA}{\imgB} &
        \methodrow{ours}{\imgA}{\imgB} &
        \methodrow{ours_prompt_rewrite}{\imgA}{\imgB} \\
        
        \methodrow{baseline}{\imgC}{\imgD} &
        \methodrow{prompt_rewrite}{\imgC}{\imgD} &
        \methodrow{ours}{\imgC}{\imgD} &
        \methodrow{ours_prompt_rewrite}{\imgC}{\imgD} \\
        
        \methodrow{baseline}{\imgE}{\imgF} &
        \methodrow{prompt_rewrite}{\imgE}{\imgF} &
        \methodrow{ours}{\imgE}{\imgF} &
        \methodrow{ours_prompt_rewrite}{\imgE}{\imgF} \\
        \bottomrule
    \end{tabularx}
    \caption{\textbf{Qualitative results for prompt rewriting.} We compare Infinity, Infinity with prompt rewriting (Infinity+P.R.), \methodname{}, and \methodname{} with prompt rewriting (\methodname{}+P.R.) under the same original prompt.}
    \label{fig:prompt_rewriting_qualitative}
    \vspace{-0.5\baselineskip}
\end{figure}

\paragraph{Prompt Rewriting.} Prompt rewriting uses an LLM to modify the input prompt to improve the generation quality and diversity. While this can introduce additional variation without much quality degradation, prompt rewriting changes the condition itself, whereas \methodname{} aims to increase diversity under the same fixed prompt. However, in practical settings where prompts are highly detailed, prompt rewriting itself has less room to introduce meaningful variation, leading to only limited changes in the generated outputs as shown in Fig.~\ref{fig:prompt_rewriting_qualitative}. Moreover, prompt rewriting is also complementary to \methodname{}, as it helps preserve image quality when combined with our method, while \methodname{} still provides substantial diversity gains. Additional discussion and quantitative results comparing \methodname{} with prompt rewriting are included in the \textbf{supplementary}.

\begingroup
\renewcommand\thefootnote{}
\footnotetext{$^\dagger$ default hyperparameter used in the original paper.}
\endgroup

\paragraph{Supplementary.}
Further details and extended results are provided in the supplementary material, including analyses of scale-travel, runtime overhead, additional metrics and comparisons, and extended qualitative and ablation results.

\section{Conclusion}
\label{sec:conclusion}
In this work, we first report a significant lack of diversity in text-conditioned VAR models, an issue that has not been addressed in prior work. Inspired by diversity enhancement techniques in diffusion models, we explore these methods and find that injection into the text embedding yields the most diverse results. However, this diversity enhancement comes at the cost of significant degradation in image quality. To address this, we propose a novel \textbf{scale-travel} latent refinement method tailored to the multi-scale nature of VAR, aiming to improve image quality while preserving diversity. Experimental results show that our two-stage method, \methodname{}, achieves the best diversity–quality Pareto frontier.

\paragraph{Limitations.} Although our scale-travel refinement technique improves the Pareto frontier, the inference cost is slightly higher than using noise injection. We also observe that the most diverse results still suffer from visual artifacts, occurring in approximately $6\%$ of the generated samples. 

\vspace{-0.5\baselineskip}
\section*{Acknowledgements.}
This work was supported by an NRF grant (RS-2026-25486000); IITP grants (RS-2024-00399817, RS-2025-25441313, RS-2025-25443318, RS-2026-25526850) funded by MSIT, Korea; the Industrial Technology Innovation Program grant (RS-2025-02317326) funded by MOTIE, Korea; the InnoCORE program of MSIT (AI Meta-Scientist, N10260110); the National Supercomputing Center with supercomputing resources and technical support (KSC-2025-CRE-0475); the Advanced GPU Utilization Support Program; and the DRB-KAIST SketchTheFuture Research Center.

%
%
\bibliographystyle{splncs04}
\bibliography{main}

\clearpage

\setcounter{section}{0}
\setcounter{figure}{0}
\setcounter{table}{0}
\renewcommand{\thesection}{\Alph{section}}
\renewcommand{\thefigure}{A\arabic{figure}}
\renewcommand{\thetable}{A\arabic{table}}
\renewcommand{\theHsection}{supp.\arabic{section}}
\renewcommand{\theHfigure}{supp.\arabic{figure}}
\renewcommand{\theHtable}{supp.\arabic{table}}

\renewcommand{\contentsname}{
DiverseVAR: Balancing Diversity and Quality of Next-Scale Visual Autoregressive Models\\[1.0\baselineskip]
{\LARGE Supplementary Material}
}

\makeatletter
\let\oldl@section\l@section
\renewcommand*\l@section[2]{
  \oldl@section{#1}{#2}
  \ifnum 1>\c@tocdepth\else\vspace{5pt}\fi
}

\let\oldl@subsection\l@subsection
\renewcommand*\l@subsection[2]{
  \oldl@subsection{#1}{#2}
  \ifnum 2>\c@tocdepth\else\vspace{5pt}\fi
}
\makeatother

\addtocontents{toc}{\protect\setcounter{tocdepth}{2}}

\begingroup
\makeatletter
\let\@mkboth\@gobbletwo
\let\l@title\@gobbletwo
\let\l@author\@gobbletwo
\makeatother
\let\clearpage\relax
\let\cleardoublepage\relax
\setcounter{tocdepth}{2}
\tableofcontents
\endgroup

\clearpage
\section{Details on Multi-Scale Encoding}
\label{sec:multiscale_encoding}
\begin{figure}[h!]
    \centering
    \includegraphics[width=\linewidth]{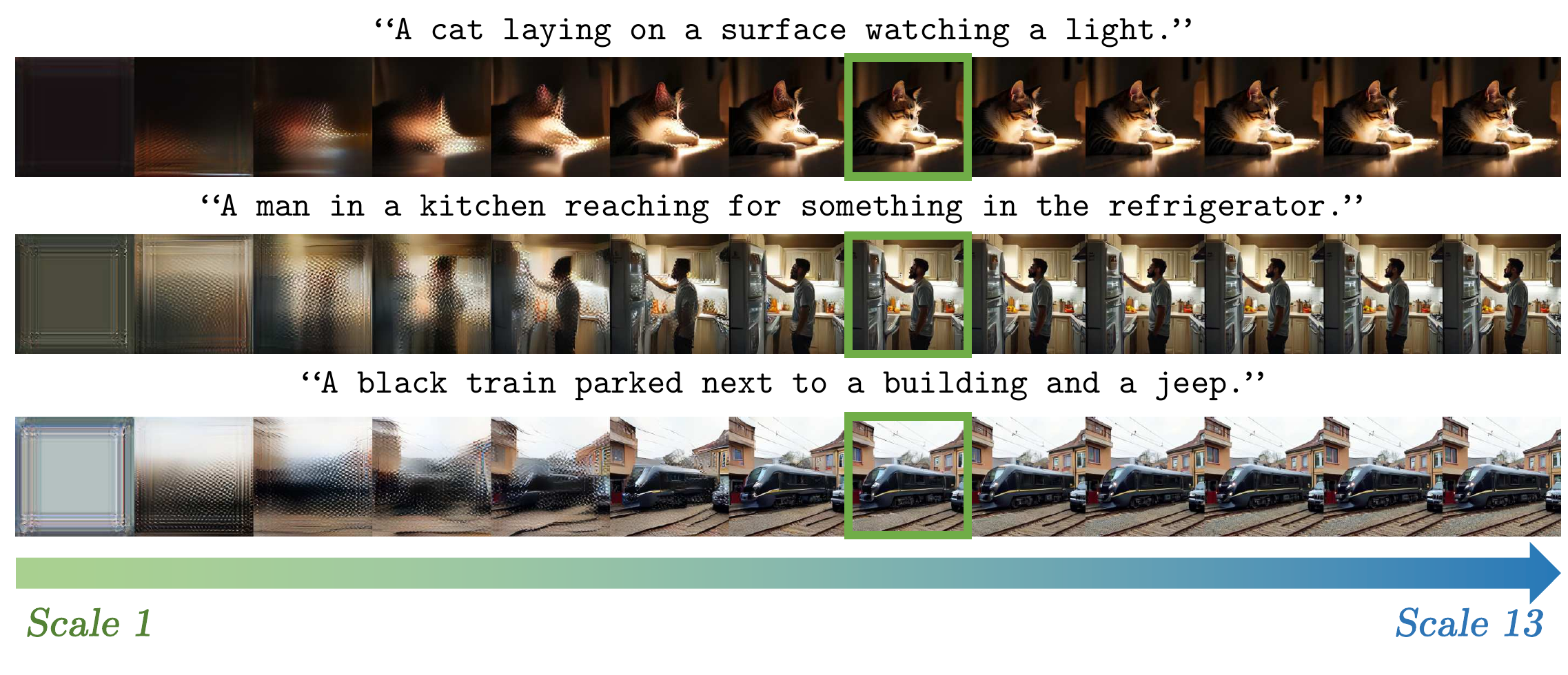}
    \vspace{-2.0\baselineskip}
    \caption{\textbf{Sampling Trajectory of VAR.} Each row displays the decoded feature map at each scale $k$ during VAR generation. The coarse high-level structure of the image is formed at earlier scales, while the fine low-level details are added at later scales.}
    \label{fig:trajectory}
\end{figure}

\vspace{-0.5\baselineskip}
As mentioned in Sec. 4.2 of the main paper, in this section, we provide additional details on the multi-scale encoding and the sampling trajectory. Fig.~\ref{fig:trajectory} displays the decoded feature map $\mathbf{Z}_k$ at each scale $k$ during VAR generation. Early scales form the coarse, high-level structure of the final image, while later scales form the fine, low-level details. Notably, by the eighth scale (\textcolor{darkgreen}{green box}), the global layout and semantic content have already emerged.

\begin{figure*}[h!]
    \centering
    \includegraphics[width=\linewidth]{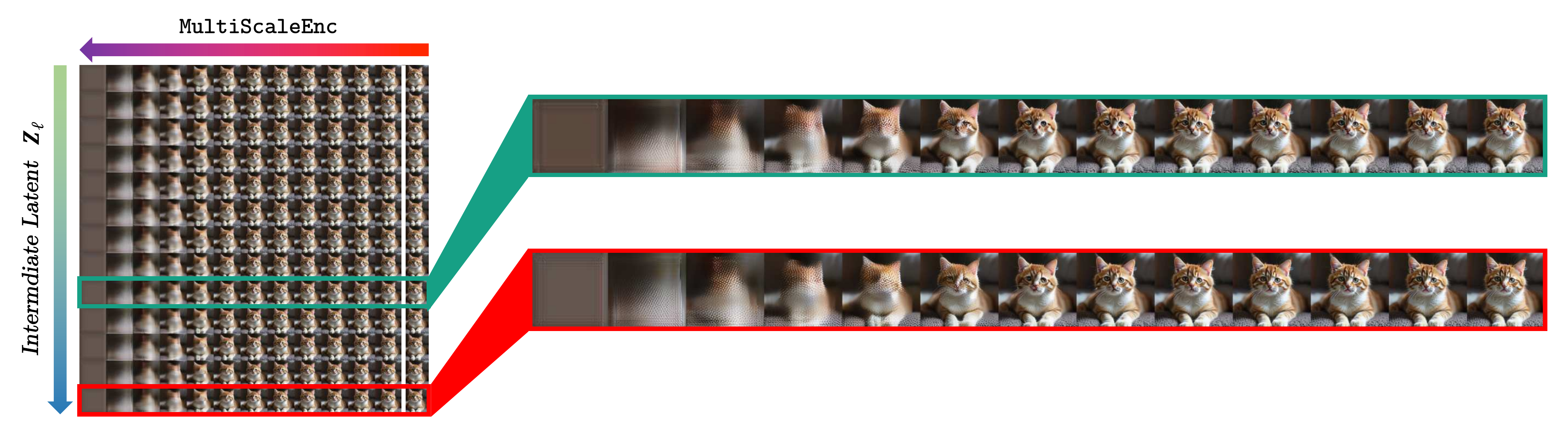}
    \vspace{-\baselineskip}
    \caption{\textbf{Multi-scale Encoding.} Multi-scale encoding applied to feature map $\mathbf{Z}_l$ for $l \in [1, K]$ (right-most column). The multi-scale encodings obtained from $\mathbf{Z}_9$ (\textcolor{teal}{cyan box}) are semantically similar to the multi-scale encodings obtained from $\mathbf{Z}_K$ (\textcolor{red}{red box}), allowing scale-travel to be performed.}
    \label{fig:multi-scale-encoding}
    \vspace{\baselineskip}
\end{figure*}

\begin{figure*}[h!]
    \centering
    \includegraphics[width=\linewidth]{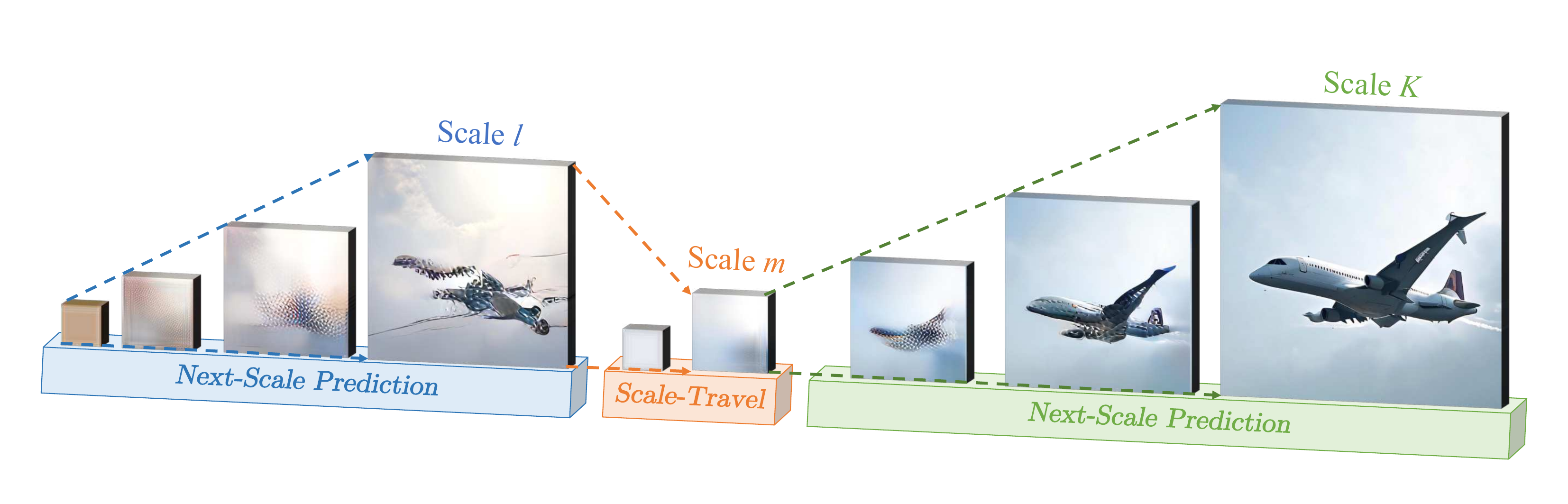}
    \vspace{-1.5\baselineskip}
    \caption{\textbf{Scale-Travel Visualization.} Scale-travel is performed by applying multi-scale encoding to the feature map at stage $l$, and resuming generation from a coarser scale $m<l$.}
    \label{fig:scale-travel-visualization}
    \vspace{\baselineskip}
\end{figure*}

Moreover, Fig.~\ref{fig:multi-scale-encoding} demonstrates the multi-scale encoding outputs of different intermediate latents $\mathbf{Z}_l$ with varying $l \in [1, K]$. In \textbf{scale-travel}, we take the multi-scale encoding of an intermediate latent (Fig.~\ref{fig:multi-scale-encoding} \textcolor{teal}{cyan box}), and run sampling initialized from the first few latents. The full scale-travel refinement strategy is visualized in Fig.~\ref{fig:scale-travel-visualization}.

\vspace{-\baselineskip}

\section{Exploration for CFG Scheduling and Condition-Annealing Noise Injection Scheduling}
\label{sec:ablation_noise_scale_schedule}
\vspace{-0.5\baselineskip}
In this section, we explore different schedulers for both CFG scheduling and noise injection techniques for diversity enhancement introduced in Sec. 4.1 of the main paper. Following previous works on CFG scheduling~\cite{wang2024analysis, Kynkaanniemi2024cfginterval}, we provide details on popular schedulers used for each variant in the below Tab.~\ref{tab:details_schedulers}, and visualize each scheduler in Fig.~\ref{fig:S3_cfg_visualize}.

\begin{figure}[h]
    \centering
    \includegraphics[width=0.6\linewidth]{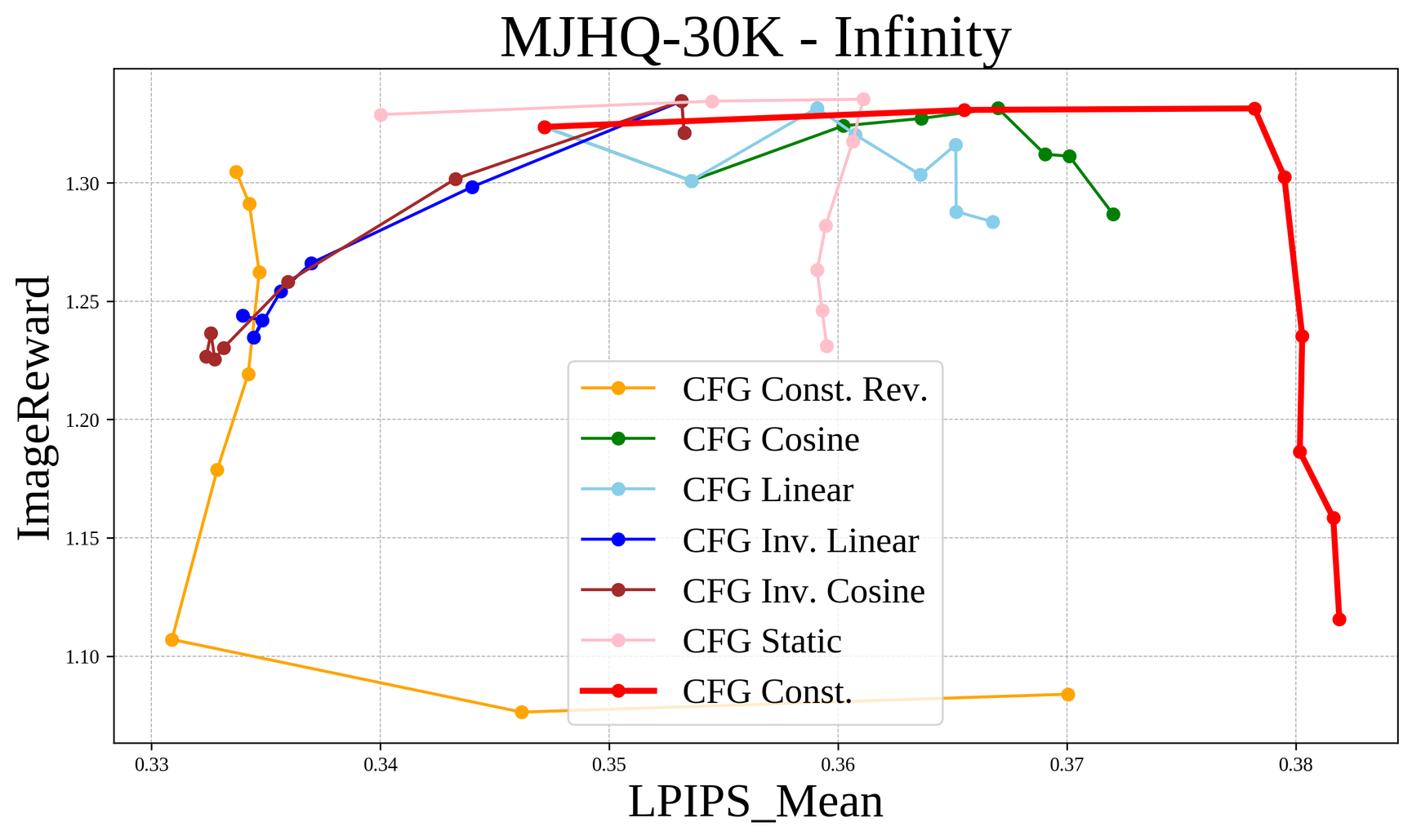}
    \vspace{-\baselineskip}
    \caption{\textbf{Pareto fronts with different CFG schedulers~\cite{Kynkaanniemi2024cfginterval, wang2024analysis}}. The plot shows the result of different schedulers applied to CFG scheduling. We find that constant CFG scheduling yields the best results.}
    \label{fig:S3_cfg}
\end{figure}

Fig.~\ref{fig:S3_cfg} shows the results for \emph{CFG-scheduling} and \emph{condition-annealing} by varying the choice of scheduling functions $\alpha(k)$, $\beta(k)$, and $\gamma(k)$. We limit the maximum value of $\alpha(k), \beta(k)$ with $\sigma$. Cosine and linear schedulers exhibit similar trends and performance on the Pareto front. The constant scheduler yields higher diversity but significantly lower image quality. Based on these results, we adopt the constant scheduler for our main experiments.

\begin{table*}[h!]
\footnotesize
\centering
\renewcommand{\arraystretch}{1.2}
\begin{tabular*}{\textwidth}{@{\extracolsep{\fill}}ccll@{}}
\toprule
\textbf{Method} & \textbf{Type} & \textbf{Variant} & \textbf{Formula / Parameters} \\
\midrule
\multirow{7}{*}{\shortstack{\textbf{CFG}\\\textbf{Scheduling}}}
  & \multirow{2}{*}{\shortstack{Piecewise\\Constant\\\cite{Kynkaanniemi2024cfginterval}}}
    & Constant 
      & $\omega_1 = 0$, $\omega_K = C$ \\
  & 
    & Constant (Inverse)
      & $\omega_1 = C$, $\omega_K = 0$ \\
\cmidrule{2-4}
  & \multirow{5}{*}{\shortstack{Interpolation\\\cite{wang2024analysis}}}
    & Cosine
      & $\gamma(k)=1 - \tfrac12\bigl(\cos(\tfrac{\pi (k-1)}{k_{\max}-1})+1\bigr)$ \\
  & 
    & Cosine (Inverse)
      & $\gamma(k)=\tfrac12\bigl(\cos(\tfrac{\pi (k-1)}{k_{\max}-1})+1\bigr)$ \\
  & 
    & Linear
      & $\gamma(k)=\tfrac{k-1}{k_{\max}-1}$ \\
  & 
    & Linear (Inverse)
      & $\gamma(k)=1 - \tfrac{k-1}{k_{\max}-1}$\\
  & 
    & Constant
      & $\gamma(k)=0.5$ \\
\midrule
\multirow{6}{*}{\shortstack{\textbf{Condition‐}\\\textbf{Annealing}}}
  & \multirow{3}{*}{\shortstack{Text\\Embedding\\\cite{sadat2023cads}}}
    & Linear
      & $\alpha(k)=\sigma\!\bigl(1-\tfrac{k-1}{k_{\max}-1}\bigr)$ \\
  &
    & Cosine
      & $\alpha(k)=\tfrac{\sigma}{2}\bigl(\cos(\tfrac{\pi (k-1)}{k_{\max}-1})+1\bigr)$ \\
  &
    & Constant
      & $\alpha(k)=\sigma$ \\
\cmidrule{2-4}
  & \multirow{3}{*}{\shortstack{\sos\ \\Token}}
    & Linear
      & $\beta(k)=\sigma\!\bigl(1-\tfrac{k-1}{k_{\max}-1}\bigr)$ \\
  &
    & Cosine
      & $\beta(k)=\tfrac{\sigma}{2}\bigl(\cos(\tfrac{\pi (k-1)}{k_{\max}-1})+1\bigr)$ \\
  &
    & Constant
      & $\beta(k)=\sigma$ \\
\bottomrule
\end{tabular*}
\vspace{0.2\baselineskip}
\caption{\textbf{Details on Diversity Enhancement Techniques.} Variants of CFG scheduling and condition‐annealing strategies~\cite{Kynkaanniemi2024cfginterval, wang2024analysis, sadat2023cads}.}
\label{tab:details_schedulers}
\vspace{-0.5\baselineskip}
\end{table*}

\vspace{-0.5\baselineskip}

\begin{figure}[h]
    \centering
    \includegraphics[width=0.9\linewidth]{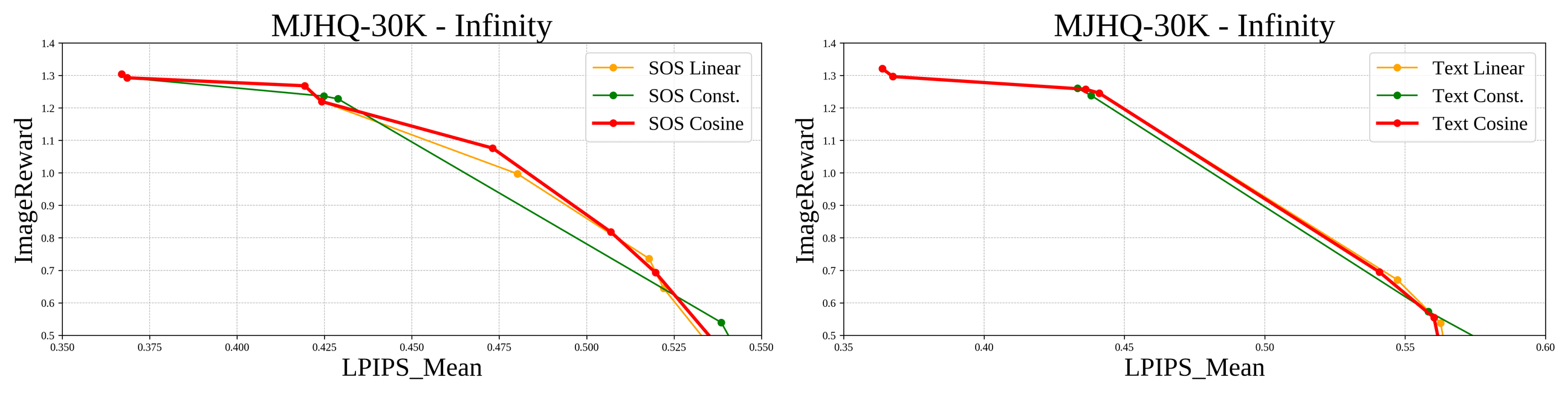}
    \vspace{-\baselineskip}
    \caption{\textbf{\sos~and Text Embedding Annealing schedulers~\cite{sadat2023cads}}. The plot shows the result of different schedulers for each of the condition-annealing methods (\sos~noise injection and Text-embedding noise injection ~\cite{sadat2023cads}). We find that cosine scheduling yields the best results.}
    \label{fig:S3_sos_cads}
\end{figure}

\vspace{-0.5\baselineskip}

\begin{figure}[h]
    \centering
    \includegraphics[width=0.9\linewidth]{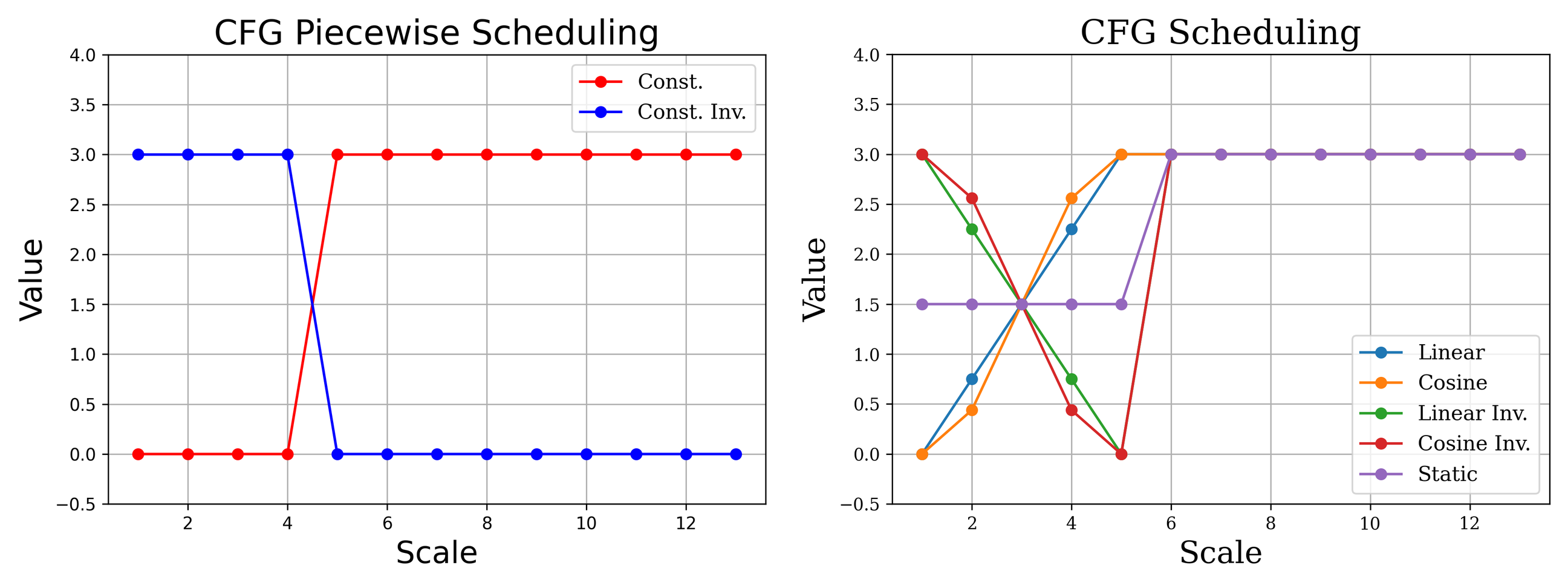}
    \vspace{-\baselineskip}
    \caption{\textbf{Visualization for various CFG schedulers with $k_{\text{max}}=5$}~\cite{Kynkaanniemi2024cfginterval, wang2024analysis}.}
    \label{fig:S3_cfg_visualize}
\end{figure}
\clearpage

\section{Comparison on Scale-Travel with various Condition-Annealing Noise Injections}
\label{sec:ablation_refinement_noise_injection}
\begin{figure}[h]
    \centering
    \includegraphics[width=\linewidth]{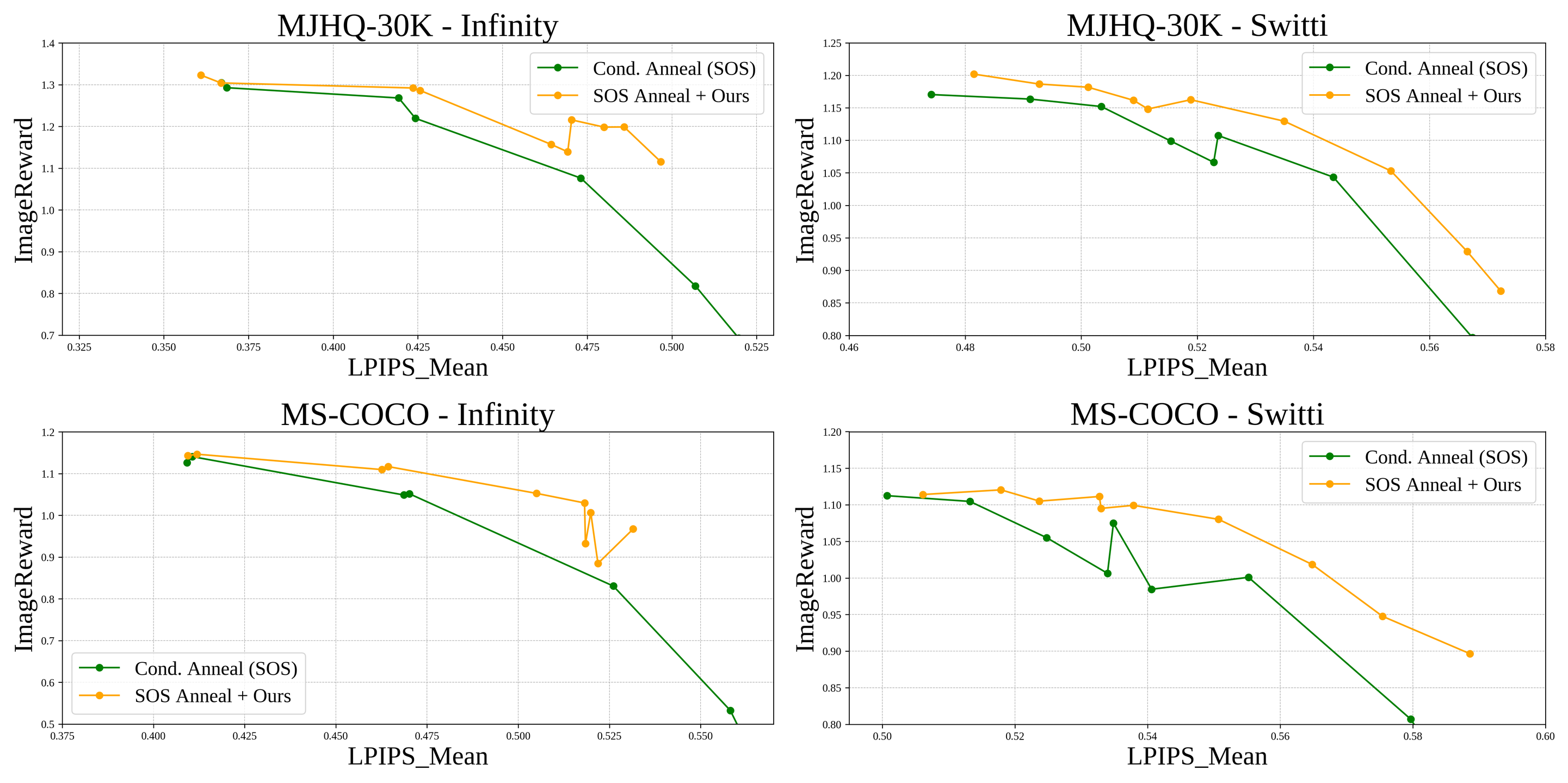}
    \caption{\textbf{Scale-Travel Applied to \sos \ Condition Annealing}. \methodname{} (SOS Anneal + Ours) consistently yields better Pareto frontiers than \sos \ Condition Annealing alone.}
    \label{fig:S4_sos}
\end{figure}

\vspace{-0.5\baselineskip}
In Fig.~\ref{fig:S4_sos}, we show Pareto fronts for the results of applying scale-travel refinement to the~\sos~condition-annealing technique. Applying scale-travel refinement to \sos~noise injection also achieves better Pareto fronts compared to the original \sos~noise injection. Fig.~\ref{fig:S4_sos} shows that using our scale-travel refinement can improve the Pareto front of any condition-annealing methods. The refinement hyperparameters used here are identical to those in the main Pareto plot in Fig. 6 in the main paper. 

\clearpage
\vspace{-\baselineskip}
\section{Details on Experiment Setup}
\label{sec:main_table_hyperparameters}
\vspace{-0.5\baselineskip}
In this section, we report the hyperparameter settings used for the main quantitative results in Tab. 1 and Tab. 5 in Sec. 5.3 of the main paper.

For Infinity~\cite{han2024infinity}, we adopt the default setting with a CFG weight of 2.0 and $\tau = 1.0$ for pure generation. Additionally, we inject noise into the text embeddings using a cosine scheduler with $k_{\text{max}} = 4$ and $\sigma = 1.0$. For refinement, we apply scale-travel from scale 8 down to scale 3.
For Switti~\cite{voronov2024switti}, we set the CFG weight to 6.0 for pure generation. Noise is injected into the text embeddings using a cosine scheduler with $k_{\text{max}} = 6$ and $\sigma = 1.0$. For refinement, scale-travel is applied from scale 8 to scale 4.

For Infinity-Star~\cite{liu2025:infinitystar}, we adopt the default setting with CFG weight of 34.0. For condition annealing, we add noise using cosine scheduler with $k_{\text{max}} = 8$ and $\sigma = 1.0$. Scale-travel is applied from scale 12 to scale 7. 
\section{Inference Time Cost Comparison}
\label{sec:inference_cost}

Tab.~\ref{tab:computation_cost} we provide a comparison of the inference time for refinement hyperparameters used in Tab.1 in Sec. 5.3 of the main paper. The measurements were conducted on an RTX 3090 GPU with 24GB of VRAM. For all settings in the table, we used a batch size of 1. The reported time costs are averaged over 100 samples. As shown, Infinity~\cite{han2024infinity} takes 2.115 seconds per inference while scale-travel takes slightly longer at 2.638 seconds per inference. Similarly, Switti~\cite{voronov2024switti} achieves an inference time of 1.642 seconds, which increases modestly to 1.892 seconds when combined with scale-travel. Notably, this corresponds to only a $15\%$ overhead, while yielding substantial improvements in the diversity–quality trade-off.

\begin{table}[h!]
\small
\centering
\begin{tabular}{lcc}
    \toprule
    \textbf{Method} & \textbf{Inference time} & \textbf{× Slower (vs. Default)} \\
    \midrule

    Infinity & 2.115s & -- \\
    + Cond. Anneal. & 2.266s & ×1.07 \\
    \rowcolor{gray!12} + Scale-Travel & 2.638s &  ×1.24 \\
    \midrule
    Switti & 1.642s & -- \\
    + Cond. Anneal. & 1.692s & ×1.03 \\
    \rowcolor{gray!12} + Scale-Travel & 1.892s & ×1.15 \\
    \bottomrule
\end{tabular}
\vspace{0.5\baselineskip}
\caption{\textbf{Inference Time Comparison.} Noise injection increases inference time because KV-caching cannot be applied during the injection process. In Scale-wise VAR, most of the computational cost is concentrated in the final stages with high-resolution feature maps, whereas \methodname{} avoids repeated inferences at these stages. Consequently, the additional overhead introduced by scale-travel remains minimal, while still providing substantial improvements in image quality.}
\label{tab:computation_cost}
\end{table}
\clearpage
\section{Additional Quantitative Results}
In this section, we present additional ablation studies and experiments to further validate the effectiveness of \methodname{}. 
Section~\ref{sec:supp_scale_travel_param_ablation} presents an ablation study on the scale-travel parameters. 
Section~\ref{sec:supp_geneval_vqa} reports the text-alignment scores, while Section~\ref{sec:supp_prompt_rewrite} compares our method with prompt rewriting. 
Section~\ref{sec:supp_additional_scale_travel} investigates the effect of applying additional scale-travel steps, and Section~\ref{sec:supp_various_models} demonstrates that \methodname{} generalizes across various model sizes and image resolutions.

\label{sec:additional_quanitatives}

\subsection{Scale-Travel Parameters for \methodname{}}
\label{sec:supp_scale_travel_param_ablation}
Scale-travel rewinds the autoregressive generation from stage $l$ to stage $m$. In Tab.~\ref{tab:extra_result_st_params}, we vary $l\in\{ 6,8,10 \}$ and $m\in\{2,3,4\}$ using MJHQ-30K~\cite{li2024mjhq} 100 prompts. We find that $m=3$ and $m=4$ yields the best diversity-quality tradeoff for Infinity and Switti, respectively.
\begin{table*}[h]
\scriptsize
\begin{tabularx}{\linewidth}{Y|YYY|YYY|YYY}
\toprule
& \multicolumn{3}{c}{$m$ = 2} & \multicolumn{3}{|c}{$m$ = 3} & \multicolumn{3}{|c}{$m$ = 4} \\
Infinity & \textbf{MPD$\uparrow$} & \textbf{Vendi$\uparrow$} & \textbf{IR$\uparrow$} & \textbf{MPD$\uparrow$} & \textbf{Vendi$\uparrow$} & \textbf{IR$\uparrow$} & \textbf{MPD$\uparrow$} & \textbf{Vendi$\uparrow$} & \textbf{IR$\uparrow$} \\
\midrule

$\ell$ = 6 & 0.385 & 4.712 & 1.312 & 0.485 & 6.108 & 1.199 & 0.556 & 7.453 & 0.724 \\
$\ell$ = 8 & 0.386 & 4.671 & 1.309 & 0.486 & 6.087 & 1.199 & 0.556 & 7.441 & 0.729 \\
$\ell$ = 10 & 0.386 & 4.668 & 1.310 & 0.486 & 6.091 & 1.184 & 0.555 & 7.430 & 0.726 \\
\midrule
Switti & \textbf{MPD$\uparrow$} & \textbf{Vendi$\uparrow$} & \textbf{IR$\uparrow$} & \textbf{MPD$\uparrow$} & \textbf{Vendi$\uparrow$} & \textbf{IR$\uparrow$} & \textbf{MPD$\uparrow$} & \textbf{Vendi$\uparrow$} & \textbf{IR$\uparrow$} \\
\midrule

$\ell$ = 6 & 0.503 & 5.737 & 1.172 & 0.542 & 6.307 & 1.115 & 0.564 & 6.695 & 0.992 \\
$\ell$ = 8 & 0.502 & 5.694 & 1.161 & 0.542 & 6.283 & 1.121 & 0.564 & 6.609 & 1.031 \\
$\ell$ = 10 & 0.502 & 5.697 & 1.179 & 0.538 & 6.226 & 1.135 & 0.563 & 6.602 & 1.004 \\
\bottomrule
\end{tabularx}

\vspace{0.5\baselineskip}
\caption{\textbf{\methodname{} with different scale-travel parameters.} Scale-travel rewinds the autoregressive generation process from stage $l$ to stage $m$. We vary $l\in\{ 6,8,10 \}$ and $m\in\{2,3,4\}$. We find that $\ell=8$ and $m=3$ yields the best diversity-quality tradeoff for Infinity and $\ell=8$ and $m=4$ yields the best diversity-quality tradeoff for Switti.}
\label{tab:extra_result_st_params}
\end{table*}

\subsection{Text-Alignment}
\label{sec:supp_geneval_vqa}
We evaluate the text-alignment of \methodname{} using GenEval~\cite{ghosh2023geneval} in Tab.~\ref{tab:extra_result_geneval} and using VQAScore~\cite{lin2024vqa} in Tab.~\ref{tab:extra_result_vqa}, following the same experimental setting as in Tab.~1 of the main paper. While \methodname{} results in slightly worse text-alignment, the decrease is not significant and can be attributed to the increased diversity: increasing the diversity leads to sampling further away from the modes of the distribution, which slightly lowers text-alignment. However, we emphasize that \methodname{} still does not significantly degrade the text-alignment according to either metric.

\subsection{Prompt Rewriting}
\label{sec:supp_prompt_rewrite}
Prompt rewriting is a simple way to improve the quality and diversity of generated images by using an LLM to augment or rephrase the input prompt. While this strategy can introduce additional variation without noticeably degrading visual quality, it differs from our setting: prompt rewriting changes the condition itself, whereas \methodname{} aims to increase diversity under the same fixed prompt.

A key desideratum in text-to-image generation is controllability, which is often achieved through detailed and specific prompts. However, as prompts become more constrained, prompt rewriting has less room to introduce meaningful variation. For example, when the prompt already specifies semantic attributes such as object type, scene, composition, lighting, and style, rewriting often changes only superficial wording, leading to limited changes in the generated outputs. This trend can be observed qualitatively in Fig. 8 of main paper, where prompt rewriting produces less variation than \methodname{}.

We quantitatively compare prompt rewriting with \methodname{} in Tab.~\ref{tab:supp_prompt_rewriting}, following the same experimental setting as Tab.~1 of the main paper. Prompt rewriting preserves image quality, but provides substantially smaller diversity gains than \methodname{}. This indicates that prompt rewriting alone does not sufficiently address the lack of per-prompt diversity in text-conditioned VAR models.

Furthermore, prompt rewriting is complementary to \methodname{}. Since prompt rewriting can make the prompt more specific, it can improve image quality when combined with our method, while \methodname{} still provides substantially larger diversity gains than prompt rewriting alone. As shown in Tab.~\ref{tab:supp_prompt_rewriting}, combining prompt rewriting with \methodname{} achieves a favorable diversity--quality trade-off, obtaining diversity comparable to \methodname{} while reducing quality degradation.

\begin{table}[h]
    \tiny 
    \setlength{\tabcolsep}{3pt}
    \centering
    \resizebox{\linewidth}{!}{
        \begin{tabular}{ l c c c c c c }
            \toprule
            & \multicolumn{3}{c}{\textbf{MJHQ-30K}} & \multicolumn{3}{c}{\textbf{MS-COCO}} \\
            \cmidrule(lr){2-4} \cmidrule(lr){5-7}
            \textbf{Method} 
            & \textbf{FID$\downarrow$} & \textbf{IR$\uparrow$} & \textbf{Vendi$\uparrow$}
            & \textbf{FID$\downarrow$} & \textbf{IR$\uparrow$} & \textbf{Vendi$\uparrow$} \\
            \midrule
            Infinity            & 19.16 & 1.22 & 4.10 & 37.37 & 1.16 & 4.41 \\
            \rowcolor{gray!12} 
            DiverseVAR          & 15.28 & 1.08 & 6.07 & 28.97 & 1.04 & 6.68 \\
            \midrule
            Infinity + P.R.     & 17.16 & 1.25 & 4.74 & 31.58 & 1.12 & 5.54 \\
            \rowcolor{gray!12} 
            DiverseVAR + P.R.   & 15.11 & 1.21 & 5.82 & 27.64 & 1.09 & 6.50 \\
            \bottomrule
        \end{tabular}
    }
    
    \vspace{0.5\baselineskip}
    \caption{\textbf{Quantitative comparison of prompt-rewrite and~\methodname{}} We compare Infinity, Infinity with prompt rewriting (P.R.), \methodname{}, and \methodname{} with prompt rewriting under the same original prompt.}
    \label{tab:supp_prompt_rewriting}
    
\end{table}

\subsection{Additional Scale-Travel Results}
\label{sec:supp_additional_scale_travel}
\vspace{-0.25\baselineskip}
We provide additional ablations in Tab.~\ref{tab:supp_repeated_scale_travel} to study the effect of applying Scale-Travel repeatedly at inference time. We follow the same experimental setting as in Tab.~1 of the main paper, using MJHQ-30K~\cite{li2024mjhq} prompts. Starting from a single Scale-Travel step, we progressively increase the number of travel steps while keeping the remaining setup unchanged. The results show that additional Scale-Travel steps can further improve generation quality, indicating that our method can trade extra computation at inference time for additional gains in quality.

\subsection{Applying to Other Model Size and Resolutions}
\label{sec:supp_various_models}
We evaluate scalability in Tab.~\ref{tab:supp-various-models} on Infinity-2B/8B across multiple resolutions, following the same experimental setting as in Tab.~1 of the main paper with MJHQ-30K~\cite{li2024mjhq} prompts. The results show that our method exhibits a consistent trend across model size and image resolutions: it increases diversity while keeping the accompanying quality drop modest.

\begin{table*}[h]
\scriptsize
\centering
\begin{tabularx}{\linewidth}{>{\centering\arraybackslash}m{0.15\textwidth}|YY>{\centering\arraybackslash}m{0.1\textwidth}YYY>{\centering\arraybackslash}m{0.12\textwidth}}
\toprule
& Overall & \makecell{Single\\Object} & Two Object & Counting & Colors & Position & \makecell{Color\\Attribution} \\
\midrule
Infinity & 0.686 & 0.997 & 0.785 & 0.622 & 0.854 & 0.258 & 0.603 \\
\rowcolor{gray!12} \methodname{} & 0.662 & 0.981 & 0.742 & 0.622 & 0.835 & 0.263 & 0.528 \\
\midrule
Switti & 0.615 & 0.997 & 0.773 & 0.544 & 0.894 & 0.118 & 0.368 \\

\rowcolor{gray!12} \methodname{} & 0.586 & 0.978 & 0.722 & 0.494 & 0.859 & 0.128 & 0.333 \\
\bottomrule
\end{tabularx}
\vspace{0.5\baselineskip}
\caption{\textbf{GenEval score of \methodname{}.} \methodname{} does not significantly degrade text-alignment.}
\label{tab:extra_result_geneval}
\end{table*}
\begin{table}[H]
\small
\centering
\begin{tabularx}{\textwidth}{Y|YY}
\toprule
& \textbf{MJHQ-30K~\cite{li2024mjhq}} & \textbf{MS-COCO~\cite{lin2014mscoco}} \\
\midrule
Infinity & 0.670 & 0.869 \\
\rowcolor{gray!12} \methodname{} & 0.663 & 0.855 \\
\midrule
Switti & 0.682 & 0.875 \\
\rowcolor{gray!12} \methodname{} & 0.671 & 0.866 \\
\bottomrule
\end{tabularx}

\vspace{0.5\baselineskip}
\caption{\textbf{VQAScore of \methodname{}.} \methodname{} does not significantly degrade the VQAScore compared to the baseline VAR models.}
\label{tab:extra_result_vqa}
\end{table}

\begin{table}[!h]
    \scriptsize
    \setlength{\tabcolsep}{2pt}
    \renewcommand{\arraystretch}{1.1}
    \centering
    \begin{tabularx}{\textwidth}{ @{} l YYY @{} }
        \toprule
        \textbf{Method} & \textbf{FID$\downarrow$} & \textbf{IR$\uparrow$} & \textbf{MPD$_{\text{L}}$ $\uparrow$} \\
        \midrule
        Infinity & 19.16 & 1.22 & 0.33 \\
        \methodname{} & 15.28 & 1.08 & 0.48 \\
        \midrule
        \rowcolor{gray!12} + Scale-Travel ($\times$2) & 15.89 & 1.12 & 0.47 \\
        \rowcolor{gray!12} + Scale-Travel ($\times$3) & 16.11 & 1.13 & 0.46 \\
        \rowcolor{gray!12} + Scale-Travel ($\times$4) & 16.19 & 1.14 & 0.46 \\
        \bottomrule
    \end{tabularx}
    
    \vspace{0.5\baselineskip}  
    \caption{\textbf{Effect of multiple Scale-Travel Steps}}
    \label{tab:supp_repeated_scale_travel}
\end{table}

\begin{table}[!h]
    
    \scriptsize

    \centering

    \begin{tabularx}{\linewidth}{l Y Y Y}
        \toprule
            \textbf{Method} & \textbf{FID$\downarrow$} & \textbf{IR$\uparrow$} & \textbf{Vendi$\uparrow$} \\
        \midrule
        Inf-8B (512,512)       & 19.49 & 1.29 & 4.51 \\
        \rowcolor{gray!12} + \methodname{}     & 15.21 & 0.81 & 7.75 \\
        \midrule
        Inf-2B (1440,720)     & 21.40 & 1.03 & 4.08 \\
        \rowcolor{gray!12} + \methodname{}      & 18.54 & 0.88 & 5.84 \\
        \midrule
        Inf-2B (720,1440)     & 20.74 & 0.85 & 4.12 \\
        \rowcolor{gray!12} + \methodname{}      & 18.78 & 0.64 & 5.71 \\
        \bottomrule
    \end{tabularx}
    
    \vspace{0.5\baselineskip}
    \caption{\textbf{Scalability Analysis.} \methodname{} consistently improves the diversity-quality tradeoff across model sizes and resolutions.}
    \label{tab:supp-various-models}
\end{table}
\clearpage
\section{Additional Video Quantitative Results}
\label{sec:additional_video_quanitatives}
In this section, we provide additional quantitative results for video generation using InfinityStar~\cite{liu2025:infinitystar} and \methodname.
We provide additional results for video quality on each VBench~\cite{huang2024vbench} and VBench2.0~\cite{zheng2025:vbench2} dimension in Tab.~\ref{tab:video-quantitative-all} and Tab.~\ref{tab:capability_eval}, respectively. As shown, condition-annealing degrades all video quality metrics whereas \methodname{} achieves comparable scores to the base model across all VBench and VBench2.0 video quality metrics.

\begin{table}[!th]

\scriptsize
\centering
\begin{tabularx}{\textwidth}{@{\extracolsep{\fill}}l Y >{\centering\arraybackslash}m{0.12\textwidth} YYYYY}
    \toprule
     \multirow{2}{*}{\textbf{Method}} & \multirow{2}{*}{\makecell{Aesthetic\\Quality}} & \multirow{2}{*}{\makecell{Imaging\\Quality}} & \multirow{2}{*}{\makecell{Background\\Consistency}} & \multirow{2}{*}{\makecell{Subject\\Consistency}} \\ \\
    \midrule
    InfinityStar~\cite{liu2025:infinitystar} & 0.571 & 0.623 & 0.952 & 0.923 \\
    \, + Cond. Anneal. & 0.559 & 0.621 & 0.934 & 0.890 \\
    \rowcolor{gray!12} \, + Scale-Travel & 0.569 & 0.637 & 0.945 & 0.915 \\
    \bottomrule
\end{tabularx}

\vspace{0.5\baselineskip}
\caption{\textbf{Quantitative comparisons for video quality using InfinityStar~\cite{liu2025:infinitystar}.} 
We evaluate all methods using VBench~\cite{huang2024vbench} on InfinityStar~\cite{liu2025:infinitystar}, a text-to-video VAR. Condition annealing degrades the video quality and consistency. 
In contrast, \methodname's scale-travel refinement achieves comparable performance with InfinityStar.}
\label{tab:video-quantitative-all}
\end{table}
\vspace{-0.75\baselineskip}
\begin{table*}[!th]
\centering
\scriptsize

\begin{tabularx}{\textwidth}{m{0.19\textwidth} YYY >{\centering\arraybackslash}m{0.12\textwidth} YY}
\toprule
\multirow{2}{*}{Model} & \multirow{2}{*}{\makecell{Complex\\Landscape}} & \multirow{2}{*}{\makecell{Camera\\Motion}} & \multirow{2}{*}{\makecell{Complex\\Plot}} & \multirow{2}{*}{Composition} & \multirow{2}{*}{Diversity} & \multirow{2}{*}{\makecell{Dynamic\\Attribute}} \\ \\
\midrule
InfinityStar~\cite{liu2025:infinitystar} & 0.193 & 0.713 & 0.145 & 0.603 & 0.270 & 0.249 \\
\, + Cond. Anneal. & 0.149 & 0.682 & 0.119 & 0.571 & 0.627 & 0.275 \\
\rowcolor{gray!12} \, + Scale-Travel & 0.180 & 0.698 & 0.130 & 0.576 & 0.487 & 0.238 \\
\bottomrule
\end{tabularx}

\vspace{0.6em}

\begin{tabularx}{\textwidth}{@{\extracolsep{\fill}} l >{\centering\arraybackslash}m{0.12\textwidth} YYYYY}
\toprule
\multirow{3}{*}{Model} & \multirow{3}{*}{\makecell{Dynamic\\Spatial\\Relationship}} & \multirow{3}{*}{\makecell{Human\\Anatomy}} & \multirow{3}{*}{\makecell{Human\\Clothes}} & \multirow{3}{*}{\makecell{Human\\Identity}} & \multirow{3}{*}{\makecell{Human\\Interaction}} & \multirow{3}{*}{\makecell{Instance\\Preservation}} \\ \\ \\
\midrule
InfinityStar~\cite{liu2025:infinitystar} & 0.362 & 0.797 & 0.903 & 0.609 & 0.787 & 0.842 \\
\, + Cond. Anneal. & 0.319 & 0.771 & 0.907 & 0.568 & 0.723 & 0.795 \\
\rowcolor{gray!12} \, + Scale-Travel & 0.357 & 0.795 & 0.872 & 0.618 & 0.743 & 0.795 \\
\bottomrule
\end{tabularx}

\vspace{0.6em}

\begin{tabularx}{\textwidth}{@{\extracolsep{\fill}}l >{\centering\arraybackslash}m{0.1\textwidth} >{\centering\arraybackslash}m{0.1\textwidth} >{\centering\arraybackslash}m{0.09\textwidth} >{\centering\arraybackslash}m{0.11\textwidth} >{\centering\arraybackslash}m{0.11\textwidth} Y}
\toprule
\multirow{3}{*}{Model} & \multirow{3}{*}{Material} & \multirow{3}{*}{Mechanics} & \multirow{3}{*}{\makecell{Motion\\Order}} & \multirow{3}{*}{\makecell{Motion\\Rationality}} & \multirow{3}{*}{\makecell{Multi-View\\Consistency}} & \multirow{3}{*}{Thermotics} \\ \\ \\
\midrule
InfinityStar~\cite{liu2025:infinitystar} & 0.789 & 0.787 & 0.374 & 0.443 & 0.243 & 0.622 \\
\, + Cond. Anneal. & 0.817 & 0.770 & 0.333 & 0.420 & 0.224 & 0.596 \\
\rowcolor{gray!12} \, + Scale-Travel & 0.818 & 0.771 & 0.401 & 0.466 & 0.264 & 0.621 \\
\bottomrule
\end{tabularx}

\vspace{0.5\baselineskip}
\caption{\textbf{Quantitative comparisons for Intrinsic Faithfulness Using VBench2.0~\cite{zheng2025:vbench2}.} We evaluate all methods using VBench2.0~\cite{zheng2025:vbench2}, which measures the intrinsic faithfulness of each video. \methodname{} outperforms condition-annealing in 14 out of 18 VBench2.0 dimensions.}
\label{tab:capability_eval}
\end{table*}
\clearpage
\section{Class-Conditioned VAR~\cite{tian2024visual}}
In this section, we further evaluate \methodname{} on the ImageNet-based class-conditioned VAR~\cite{tian2024visual}. 
We find that class-conditioned VAR also exhibits limited diversity and employs the CFG-annealing technique. 
We evaluate the quality-diversity trade-off using FID~\cite{heusel2017fid}, Precision, and Recall~\cite{sajjadi2018precrec,kynkaanniemi2019genprecrec}, with 10 images generated for each ImageNet class.
As shown in Tab.~\ref{tab:supp_class_var}, our method improves diversity in this setting as well, consistent with the trends observed across various text-conditioned VARs in the main paper. 
In detail, we set the CFG weight to 1.5 for pure generation. Noise is injected into the class label embeddings using a cosine scheduler with $k_{\text{max}} = 3$ and $\sigma = 1.0$. For refinement, scale-travel is applied from scale 8 to scale 2.
\vspace{\baselineskip}
\begin{table}[h]
    \scriptsize
    \centering
    \scriptsize
    \begin{tabularx}{\textwidth}{l YYY}
        \toprule
        \textbf{Method} & \textbf{FID$\downarrow$} & \textbf{Prec.$\uparrow$} & \textbf{Rec.$\uparrow$} \\
        \midrule
        VAR     & 10.578 & 0.877 & 0.543 \\
        \, + Cond. Anneal.  & 10.357 & 0.674 & 0.758 \\
        \rowcolor{gray!12} \, + Scale-Travel  & 5.560 & 0.795 & 0.693 \\
        \bottomrule
    \end{tabularx}
        
    \vspace{0.5\baselineskip}
    \caption{\textbf{Results on Class-Conditioned VAR~\cite{tian2024visual}.} \methodname{} significantly improves the diversity (recall) of class-conditioned VAR while mitigating the degradation in image quality (precision). Due to this improved diversity-quality tradeoff, \methodname{} achieves better FID than VAR (without CFG annealing).}
    \label{tab:supp_class_var}
\end{table}
\clearpage
\section{Comparison with Fine-Tuning Models}
In this section, we present a quantitative comparison against recent fine-tuning-based baselines, including CSD-VAR~\cite{nguyen2025csd} and Kaleido~\cite{gu2024kaleido}. We follow each method's recommended setup as closely as possible, and report FID, image realism (IR), and Vendi (diversity) scores using MJHQ-30K~\cite{li2024mjhq} in Table~\ref{tab:supp-fine-tuning}.

\begin{itemize}
    \item {\bf CSD-VAR~\cite{nguyen2025csd}}: Following CSD-VAR, we apply textual inversion to produce stylistically diverse generations. While this increases diversity, it often introduces noticeable artifacts and degrades visual quality, especially for more complex prompts.
    \item {\bf Kaleido~\cite{gu2024kaleido}:} We follow Kaleido by sampling 100k high-resolution LAION images and constructing a latent-token dataset using Qwen2.5-VL~\cite{bai2025:qwen25vl} and SAM~\cite{kirillov2023:sam}, with the SEED tokenizer~\cite{ge2023:seedtok} for discrete visual tokens. We then train an additional text-conditioning module by fine-tuning the decoder of a T5-large~\cite{colin2020:T5} and feed its hidden states as extra conditions to the pretrained VAR. Finally, we LoRA-finetune~\cite{hu2022:lora} Infinity-2B while updating only the newly introduced parameters. This improves diversity, but our method yields a larger diversity gain while better preserving image quality.
\end{itemize}

\begin{table}[!h]
    \scriptsize

    \centering
    \vspace{-8pt}

    \begin{tabularx}{\textwidth}{ l YYY }
        \toprule
        \textbf{Method} & \textbf{FID$\downarrow$} & \textbf{IR$\uparrow$} & \textbf{Vendi$\uparrow$} \\
        \midrule
        
        \rowcolor{gray!12} \methodname{} & 15.28 & 1.08 & 6.07 \\
        \midrule
        CSD-VAR~\cite{nguyen2025csd} & 14.71 & 0.62 & 6.81 \\
        Kaleido~\cite{gu2024kaleido} & 17.04 & 0.92 & 5.88 \\

        \bottomrule
    \end{tabularx}
    
    \vspace{0.5\baselineskip}
    \caption{\textbf{Comparison with fine-tuning-based methods.} We compare \methodname{} against the fine-tuning-based methods CSD-VAR~\cite{nguyen2025csd} and Kaleido~\cite{gu2024kaleido}. Without any additional training, \methodname{} achieves significantly better image quality than CSD-VAR while attaining high diversity, and outperforms Kaleido in both quality and diversity.}
    \label{tab:supp-fine-tuning}
\end{table}
\clearpage
\section{Additional Image Qualitative Results}
\label{sec:additional_qualitatives}

\begin{figure*}[h]
    \centering
    
    \setlength{\tabcolsep}{0pt}
    \scriptsize
    \vspace{-0.25\baselineskip}
    \begin{tabularx}{0.8\linewidth}{Y Y Y | Y Y Y | Y Y Y}
        \toprule
    \multicolumn{3}{Y|}{Infinity~\cite{han2024infinity}} & \multicolumn{3}{Y|}{+ Condition Annealing.} & \multicolumn{3}{Y}{+ Scale-Travel \textbf{(Ours)}.} \\
        \midrule
        \multicolumn{9}{c}{``\texttt{\input{figures/results/qualitative_samples/mscoco/prompts/0.txt}}\unskip''} \\
        \multicolumn{9}{c}{\includegraphics[width=0.8\linewidth]{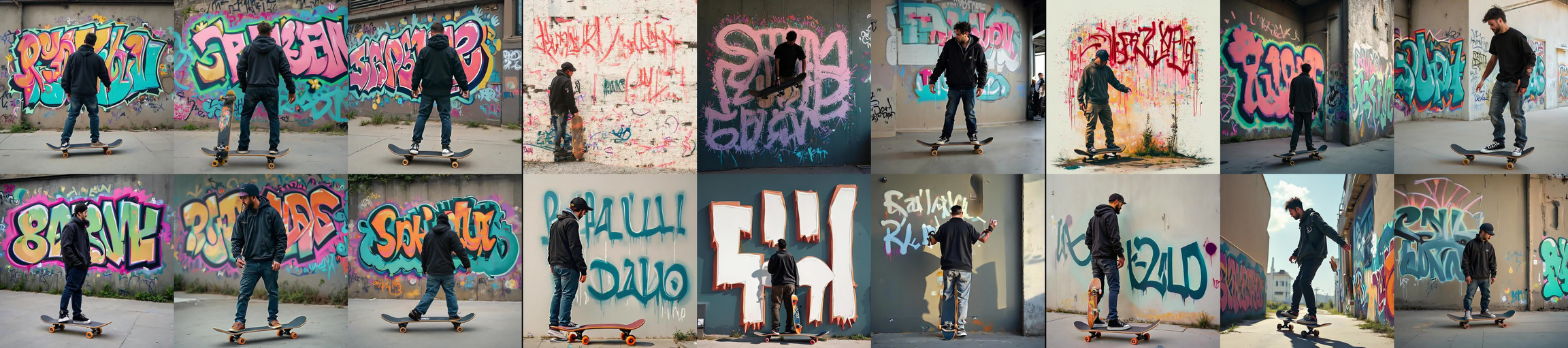}} \\
        \multicolumn{9}{c}{``\texttt{\input{figures/results/qualitative_samples/mscoco/prompts/1.txt}}\unskip''} \\
        \multicolumn{9}{c}{\includegraphics[width=0.8\linewidth]{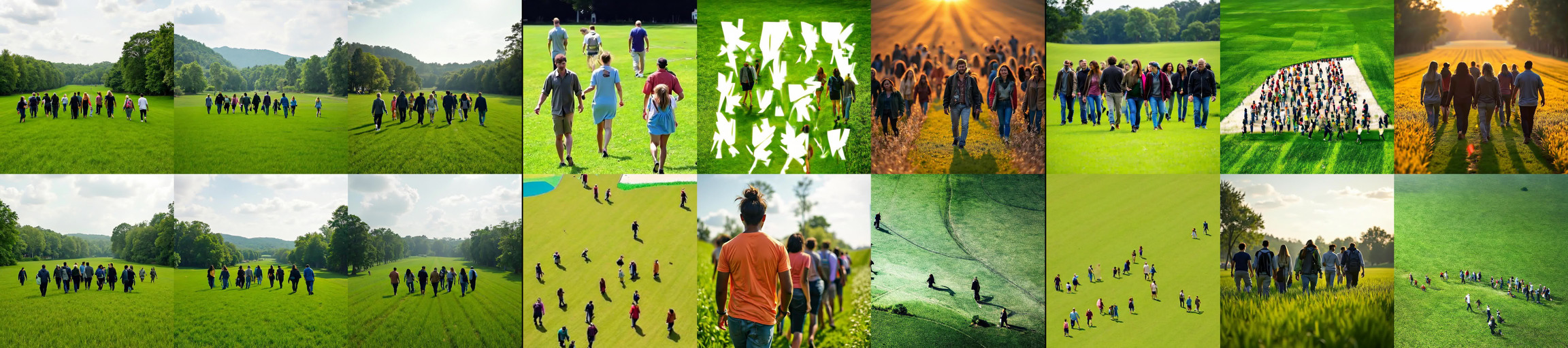}} \\
        \multicolumn{9}{c}{``\texttt{\input{figures/results/qualitative_samples/mscoco/prompts/2.txt}}\unskip''} \\
        \multicolumn{9}{c}{\includegraphics[width=0.8\linewidth]{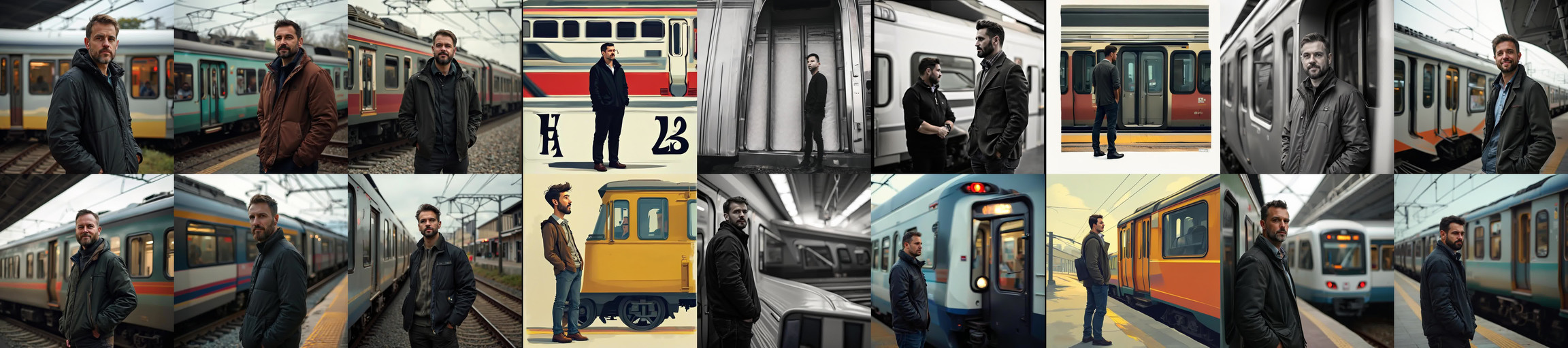}} \\
        \multicolumn{9}{c}{``\texttt{\input{figures/results/qualitative_samples/mscoco/prompts/3.txt}}\unskip''} \\
        \multicolumn{9}{c}{\includegraphics[width=0.8\linewidth]{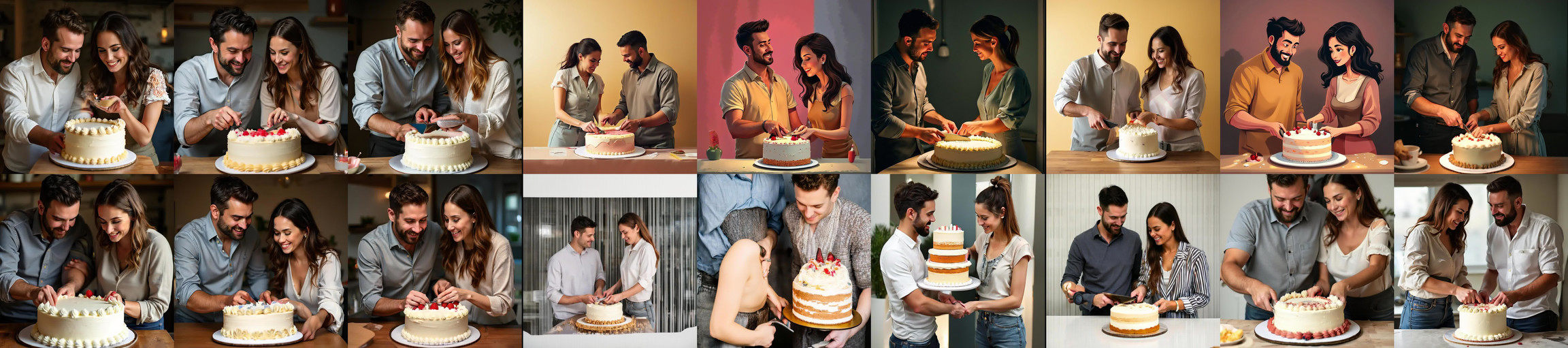}} \\
        \multicolumn{9}{c}{``\texttt{\input{figures/results/qualitative_samples/mscoco/prompts/4.txt}}\unskip''} \\
        \multicolumn{9}{c}{\includegraphics[width=0.8\linewidth]{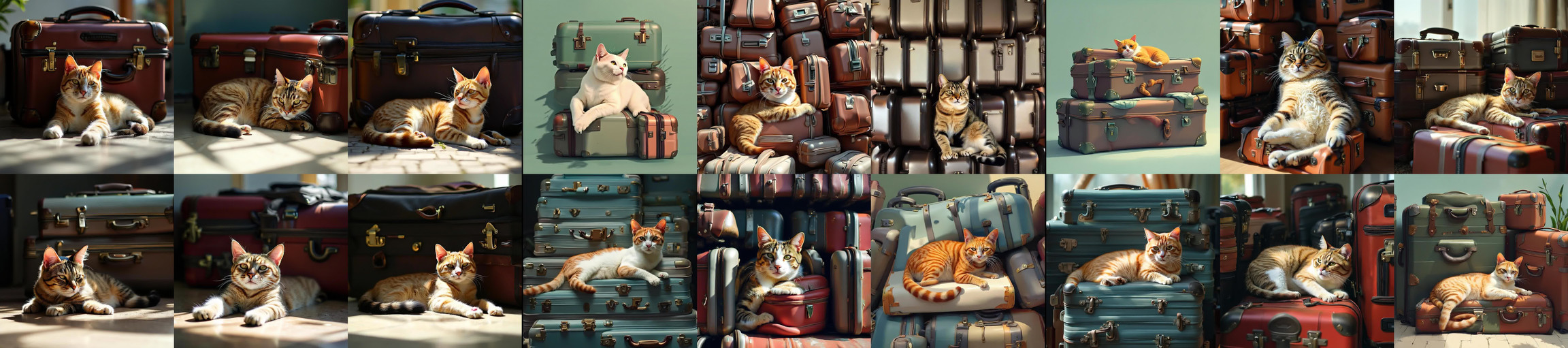}} \\
        \bottomrule
    \end{tabularx}
    \vspace{-\baselineskip}
    \caption{\textbf{Additional Qualitative Results on MS-COCO~\cite{lin2014mscoco} using Infinity~\cite{han2024infinity}.} Condition-annealing improves the diversity at the cost of visual artifacts and lower image quality. Applying Scale-Travel \textbf{(Ours)} maintains a similar degree of diversity while correcting the visual artifacts.}
    \label{fig:supp-mscoco-qualitatives}
\end{figure*}

\begin{figure*}[h]
    \centering
    \setlength{\tabcolsep}{0pt}
    \scriptsize
    \vspace{-0.25\baselineskip}
    \begin{tabularx}{0.8\linewidth}{Y Y Y | Y Y Y | Y Y Y}
        \toprule
        \multicolumn{3}{Y|}{Switti~\cite{voronov2024switti}} & \multicolumn{3}{Y|}{+ Condition Annealing.} & \multicolumn{3}{Y}{+ Scale-Travel \textbf{(Ours)}.} \\
        \midrule
        \multicolumn{9}{c}{``\texttt{\input{figures/results/qualitative_samples/mscoco_switti/prompts/0.txt}}\unskip''} \\
        \multicolumn{9}{c}{\includegraphics[width=0.8\linewidth]{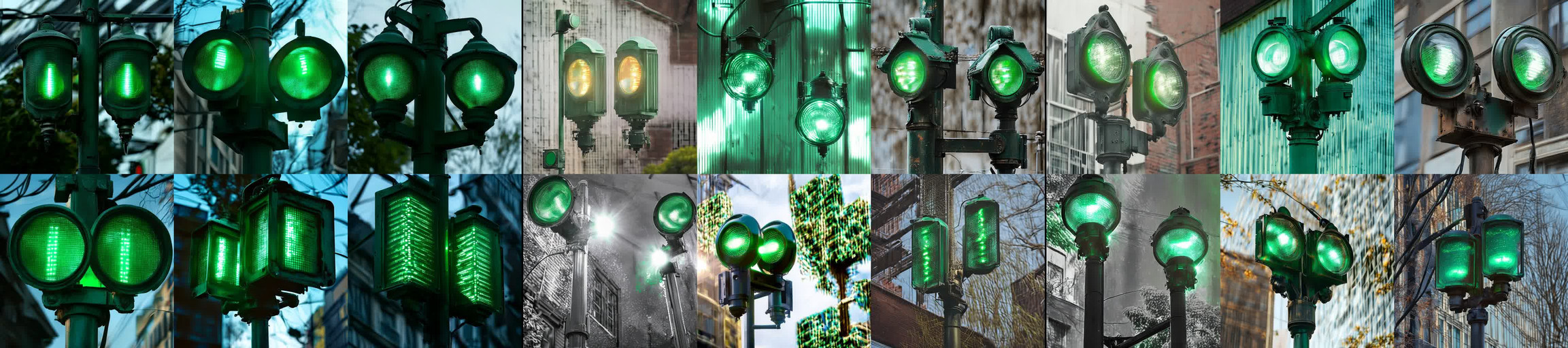}} \\
        \multicolumn{9}{c}{``\texttt{\input{figures/results/qualitative_samples/mscoco_switti/prompts/1.txt}}\unskip''} \\
        \multicolumn{9}{c}{\includegraphics[width=0.8\linewidth]{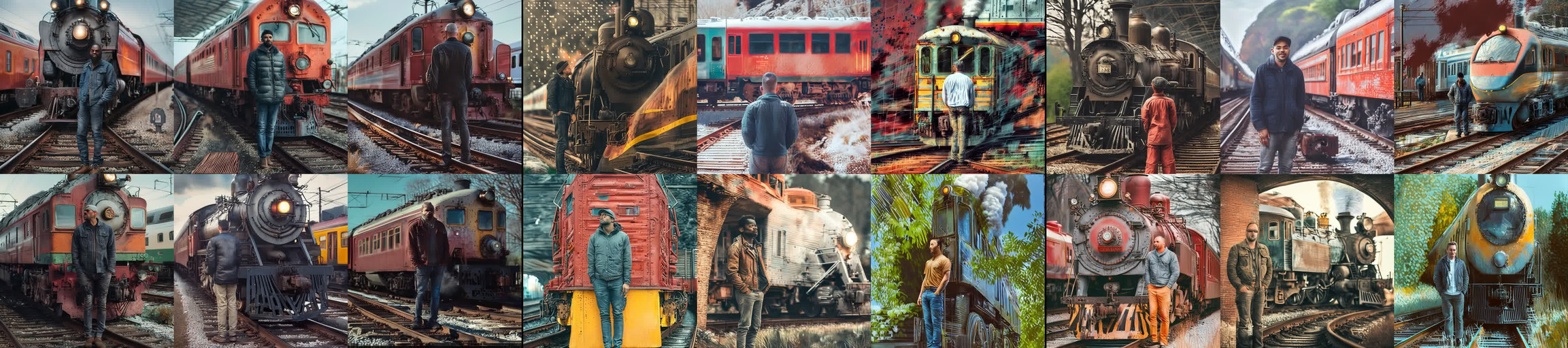}} \\
        \multicolumn{9}{c}{``\texttt{\input{figures/results/qualitative_samples/mscoco_switti/prompts/2.txt}}\unskip''} \\
        \multicolumn{9}{c}{\includegraphics[width=0.8\linewidth]{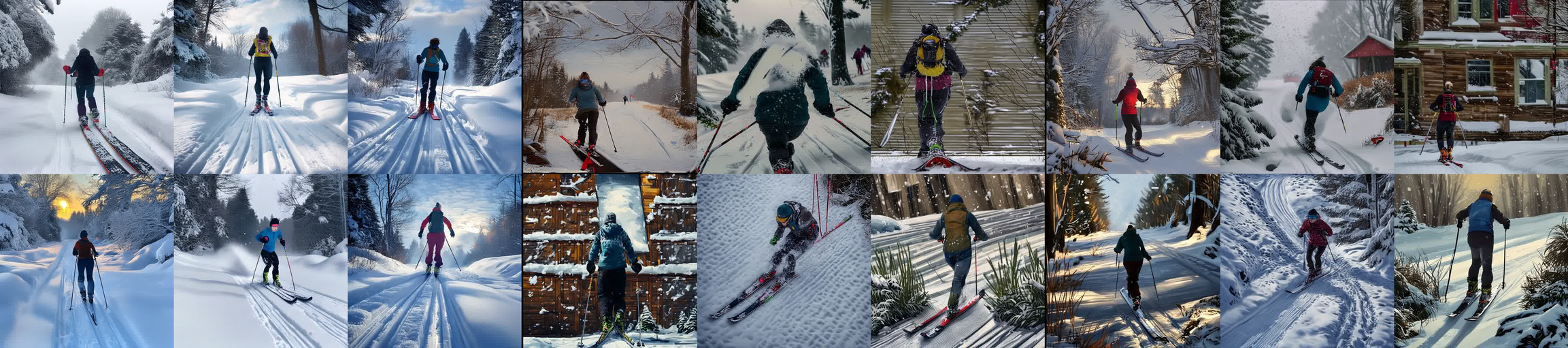}} \\
        \multicolumn{9}{c}{``\texttt{\input{figures/results/qualitative_samples/mscoco_switti/prompts/3.txt}}\unskip''} \\
        \multicolumn{9}{c}{\includegraphics[width=0.8\linewidth]{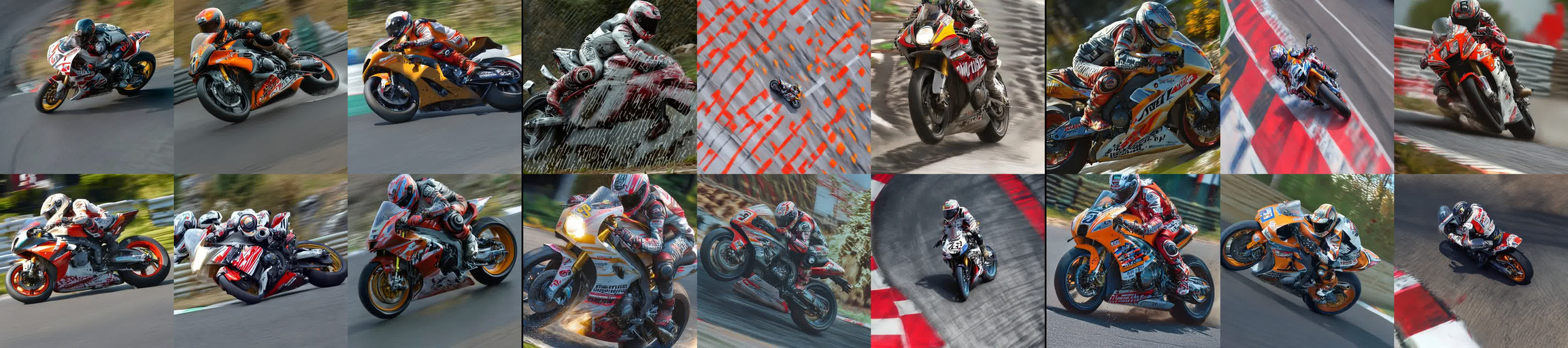}} \\
        \multicolumn{9}{c}{``\texttt{\input{figures/results/qualitative_samples/mscoco_switti/prompts/4.txt}}\unskip''} \\
        \multicolumn{9}{c}{\includegraphics[width=0.8\linewidth]{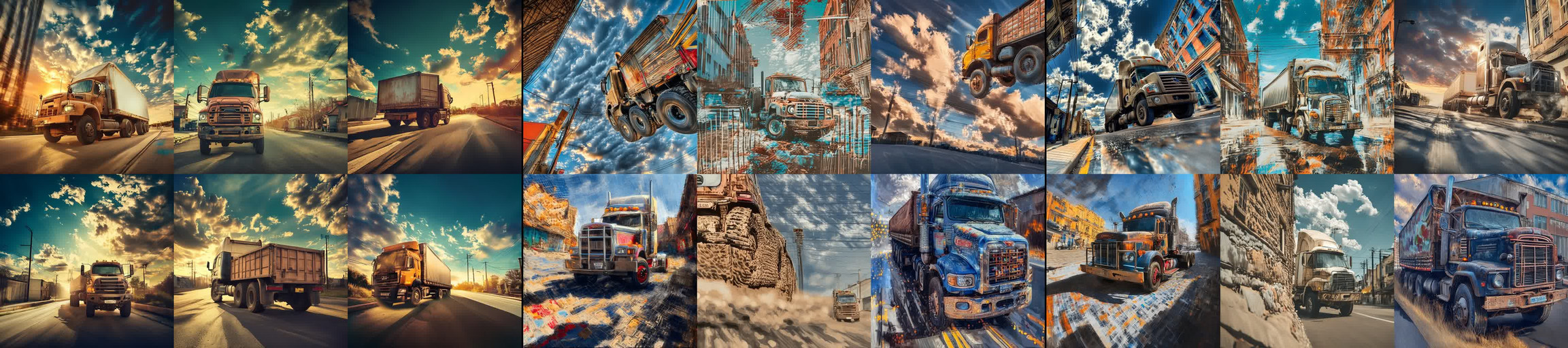}} \\
        \bottomrule
    \end{tabularx}
    \vspace{-\baselineskip}
    \caption{\textbf{Additional Qualitative Results on MS-COCO~\cite{lin2014mscoco} using Switti~\cite{voronov2024switti}.} Condition-annealing improves the diversity at the cost of visual artifacts and lower image quality. Applying Scale-Travel \textbf{(Ours)} maintains a similar degree of diversity while correcting the visual artifacts.}
    \label{fig:supp-mscoco-qualitatives-pt2}
\end{figure*}

\begin{figure*}[h]
    \centering
    
    \setlength{\tabcolsep}{0pt}
    \scriptsize
    \vspace{-0.25\baselineskip}
    \begin{tabularx}{0.8\linewidth}{Y Y Y | Y Y Y | Y Y Y}
        \toprule
        \multicolumn{3}{Y|}{Infinity~\cite{han2024infinity}} & \multicolumn{3}{Y|}{+ Condition Annealing.} & \multicolumn{3}{Y}{+ Scale-Travel \textbf{(Ours)}.} \\
        \midrule
        \multicolumn{9}{c}{\parbox{0.8\textwidth}{\centering``\texttt{\input{figures/results/qualitative_samples/mjhq/prompts/0.txt}}\unskip''}} \\
        \multicolumn{9}{c}{\includegraphics[width=0.8\linewidth]{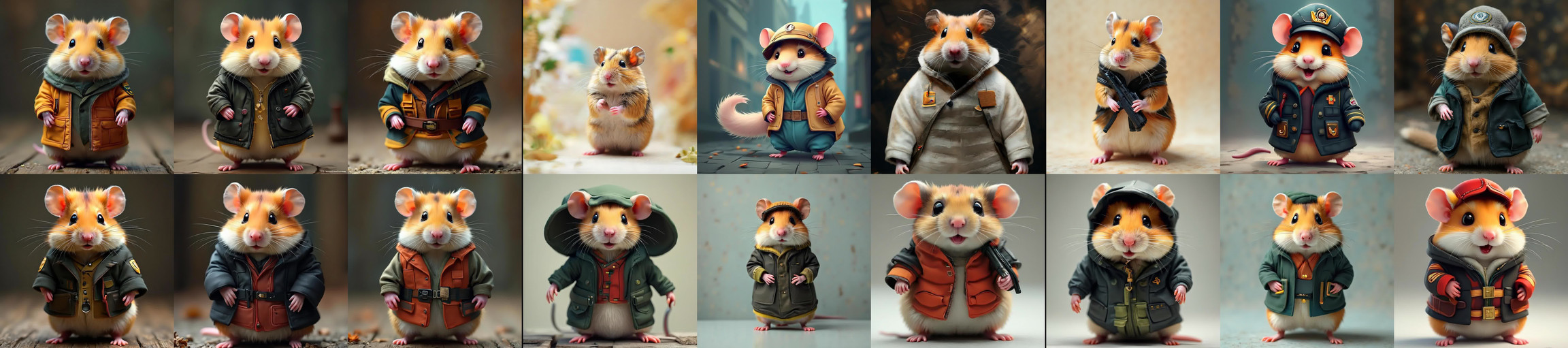}} \\
        \multicolumn{9}{c}{\parbox{0.8\textwidth}{\centering``\texttt{\input{figures/results/qualitative_samples/mjhq/prompts/1.txt}}\unskip''}} \\
        \multicolumn{9}{c}{\includegraphics[width=0.8\linewidth]{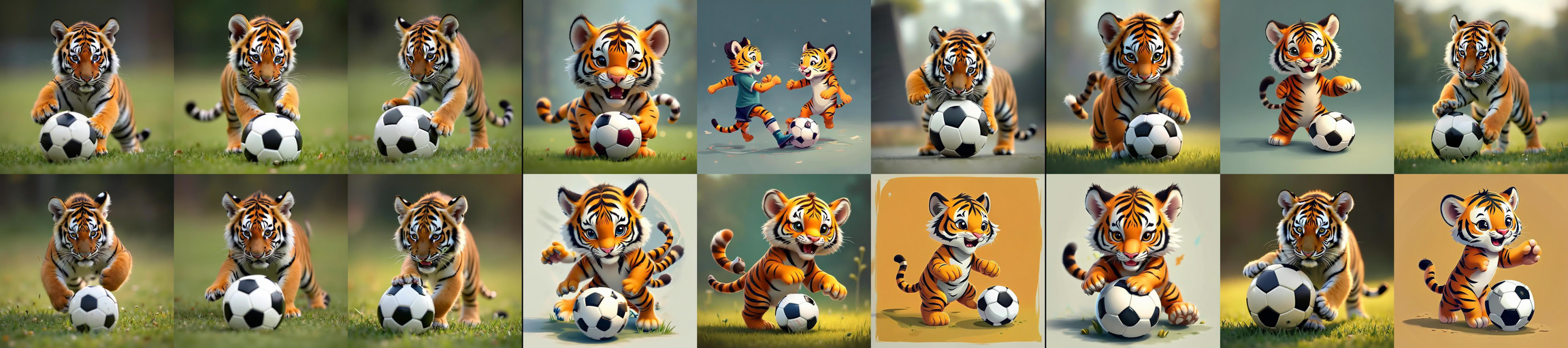}} \\
        \multicolumn{9}{c}{\parbox{0.8\textwidth}{\centering``\texttt{\input{figures/results/qualitative_samples/mjhq/prompts/2.txt}}\unskip''}} \\
        \multicolumn{9}{c}{\includegraphics[width=0.8\linewidth]{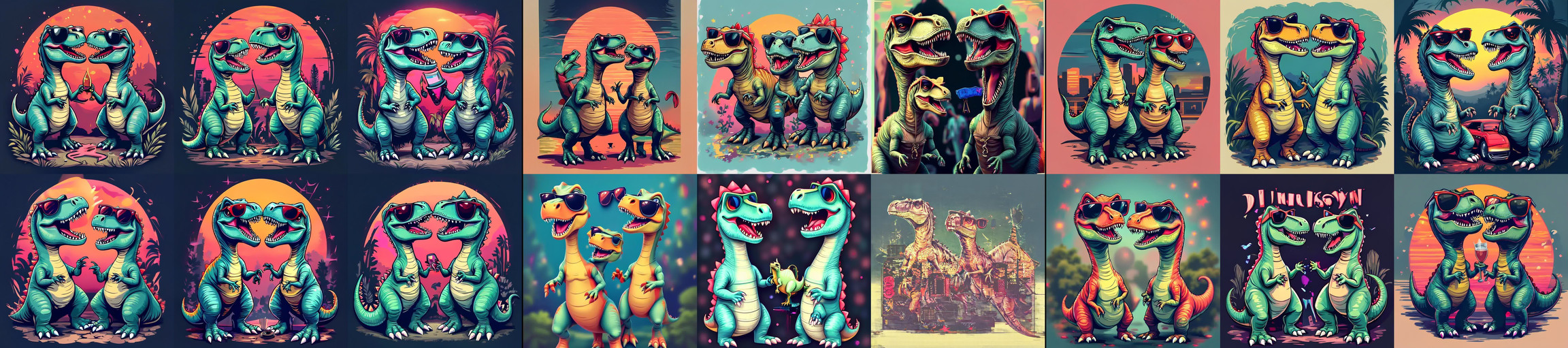}} \\
        \multicolumn{9}{c}{\parbox{0.8\textwidth}{\centering``\texttt{\input{figures/results/qualitative_samples/mjhq/prompts/3.txt}}\unskip''}} \\
        \multicolumn{9}{c}{\includegraphics[width=0.8\linewidth]{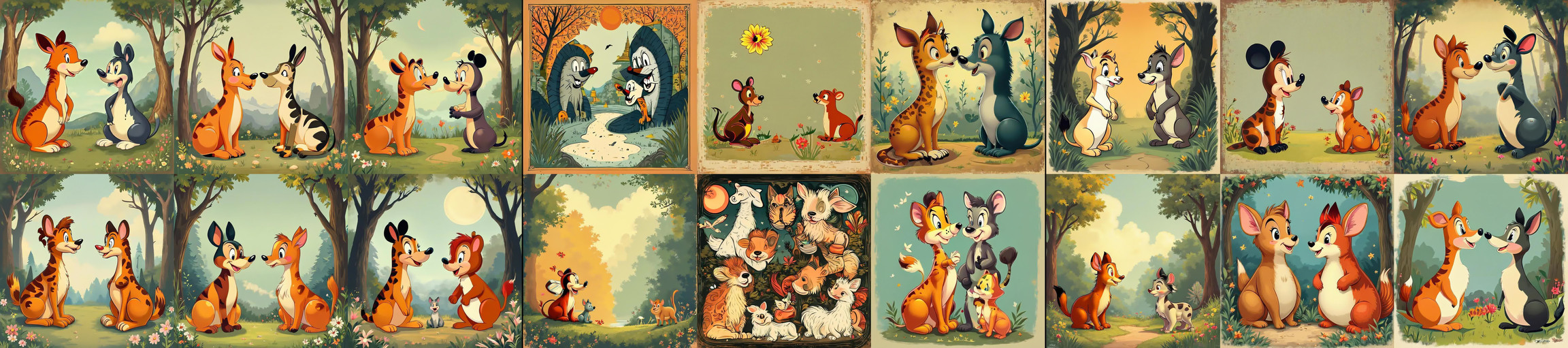}} \\
        \multicolumn{9}{c}{\parbox{0.8\textwidth}{\centering``\texttt{\input{figures/results/qualitative_samples/mjhq/prompts/4.txt}}\unskip''}} \\
        \multicolumn{9}{c}{\includegraphics[width=0.8\linewidth]{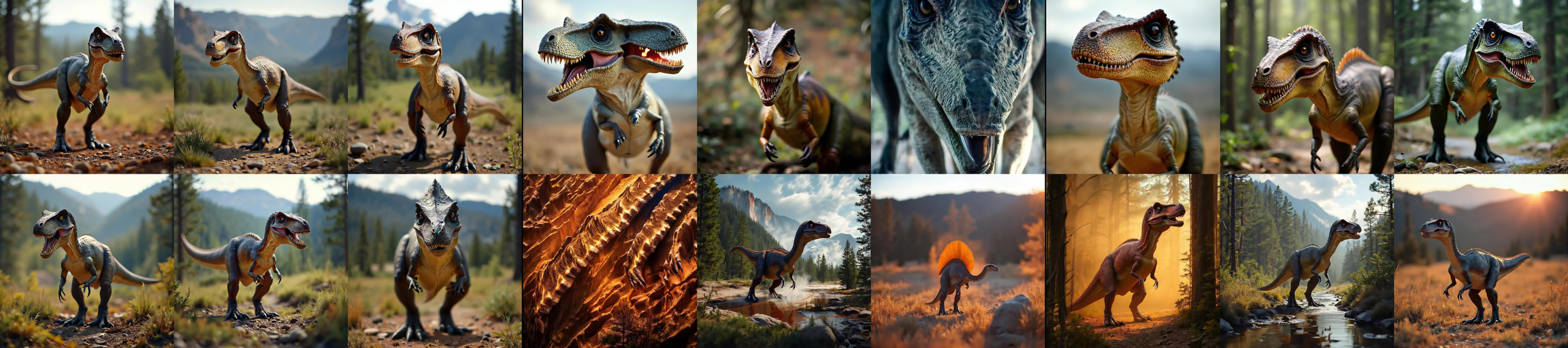}} \\
        \bottomrule
    \end{tabularx}
    \vspace{-\baselineskip}
    \caption{\textbf{Additional Qualitative Results. MJHQ-30K~\cite{li2024mjhq} using Infinity~\cite{han2024infinity}.} Condition-annealing improves the diversity at the cost of visual artifacts and lower image quality. Applying Scale-Travel \textbf{(Ours)} maintains a similar degree of diversity while correcting the visual artifacts.}
    \label{fig:supp-mjhq-qualitatives}
    
\end{figure*}

\begin{figure*}[h]
    \centering
    \setlength{\tabcolsep}{0pt}
    \scriptsize
    \vspace{-0.25\baselineskip}
    \begin{tabularx}{0.8\linewidth}{Y Y Y | Y Y Y | Y Y Y}
        \toprule
        \multicolumn{3}{Y|}{Switti~\cite{voronov2024switti}} & \multicolumn{3}{Y|}{+ Condition Annealing.} & \multicolumn{3}{Y}{+ Scale-Travel \textbf{(Ours)}.} \\
        \midrule
        \multicolumn{9}{c}{\parbox{0.8\textwidth}{\centering``\texttt{\input{figures/results/qualitative_samples/mjhq_switti/prompts/0.txt}}\unskip''}} \\
        \multicolumn{9}{c}{\includegraphics[width=0.8\linewidth]{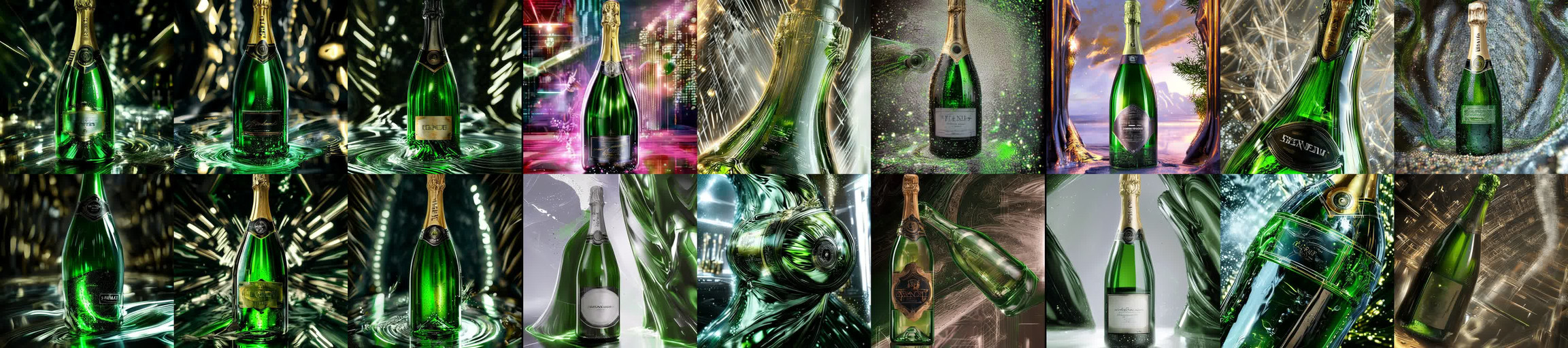}} \\
        \multicolumn{9}{c}{\parbox{0.8\textwidth}{\centering``\texttt{\input{figures/results/qualitative_samples/mjhq_switti/prompts/2.txt}}\unskip''}} \\
        \multicolumn{9}{c}{\includegraphics[width=0.8\linewidth]{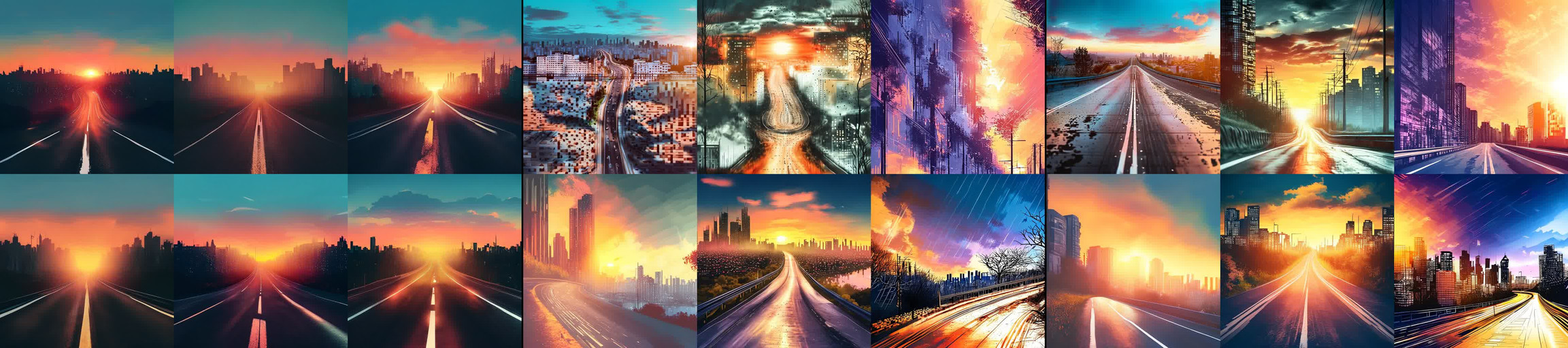}} \\
        \multicolumn{9}{c}{\parbox{0.8\textwidth}{\centering``\texttt{\input{figures/results/qualitative_samples/mjhq_switti/prompts/3.txt}}\unskip''}} \\
        \multicolumn{9}{c}{\includegraphics[width=0.8\linewidth]{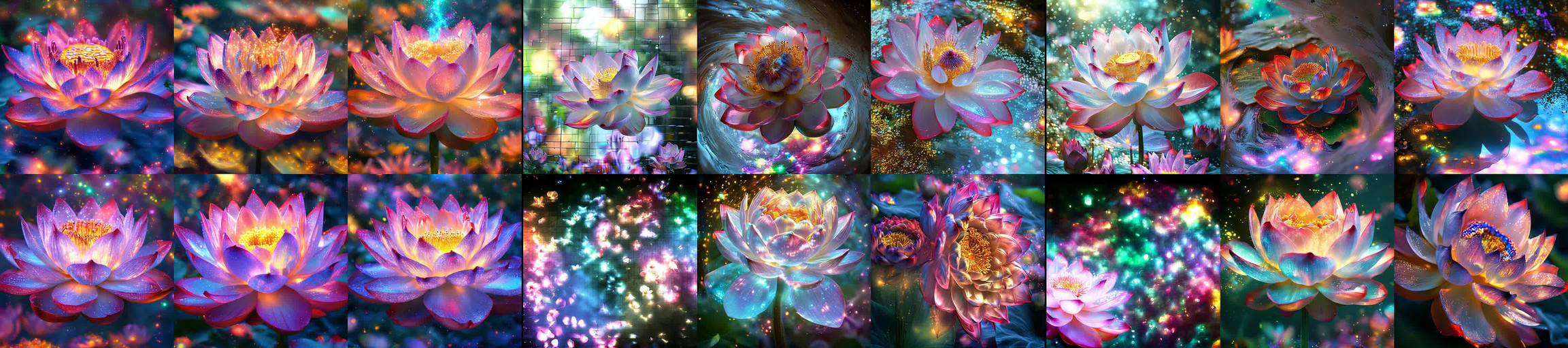}} \\
        \multicolumn{9}{c}{\parbox{0.8\textwidth}{\centering``\texttt{\input{figures/results/qualitative_samples/mjhq_switti/prompts/4.txt}}\unskip''}} \\
        \multicolumn{9}{c}{\includegraphics[width=0.8\linewidth]{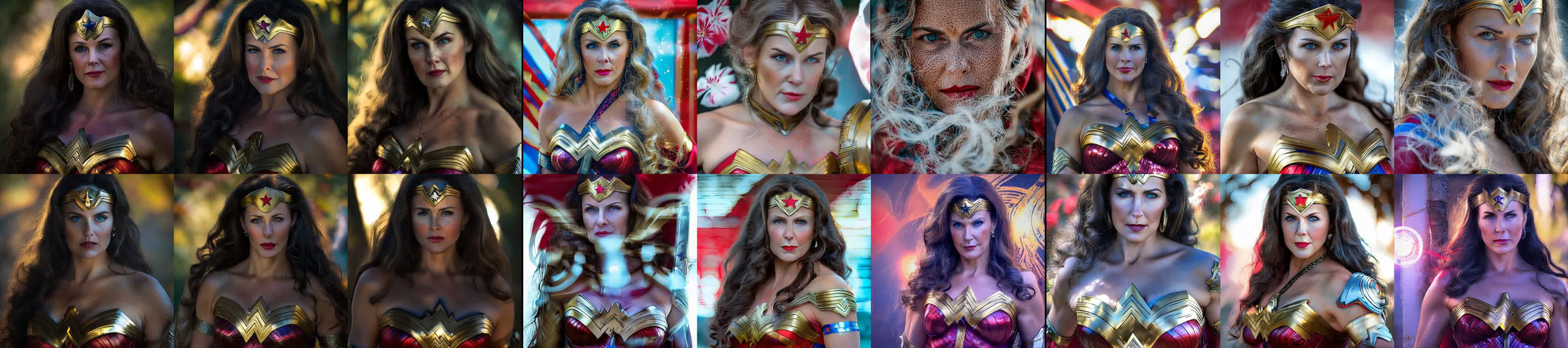}} \\
        \bottomrule
    \end{tabularx}
    \vspace{-\baselineskip}
    \caption{\textbf{Additional Qualitative Results. MJHQ-30K~\cite{li2024mjhq} using Switti~\cite{voronov2024switti} (Cont'd).} Condition-annealing improves the diversity at the cost of visual artifacts and lower image quality. Applying Scale-Travel \textbf{(Ours)} maintains a similar degree of diversity while correcting the visual artifacts.}
    \label{fig:supp-mjhq-qualitatives-pt2}
\end{figure*}

We provide additional qualitative results for MS-COCO~\cite{lin2014mscoco} in Fig.~\ref{fig:supp-mscoco-qualitatives} and Fig.~\ref{fig:supp-mscoco-qualitatives-pt2} and for MJHQ-30k~\cite{li2024mjhq} in Fig.~\ref{fig:supp-mjhq-qualitatives} and Fig.~\ref{fig:supp-mjhq-qualitatives-pt2}. Condition-annealing increases the diversity at the expense of image quality. Adding our scale-travel refinement technique strikes the balance between diversity and image quality.

\clearpage
\section{Additional Video Qualitative Results}
\label{sec:additional_video_qualitatives}

We provide additional qualitative results for InfinityStar~\cite{liu2025:infinitystar} in Fig.~\ref{fig:video_qualitative_ap1} and Fig.~\ref{fig:video_qualitative_ap2}. Condition-annealing increases the diversity at the expense of image quality. Adding our scale-travel refinement technique strikes the balance between diversity and image quality. For generation, we employ augmented prompts to obtain higher-quality outputs.

\newcommand{\VideoComparisonPanel}[1]{
\begin{minipage}{0.49\textwidth}
\centering
\newcommand{\vidframe}[3]{figures/results/video_samples/##1/#1/##2_##3.jpg}
\newcommand{\vidprompt}{figures/results/video_samples/prompts/#1.txt}
\begin{tabular}{>{\centering\arraybackslash}p{0.9em} cccc >{\centering\arraybackslash}p{0.5em}}
{} & \scriptsize Frame 1 & \scriptsize Frame 26 & \scriptsize Frame 51 & \scriptsize Frame 75 & {} \\
\multirow{3}{*}[1.5ex]{\rotatebox{90}{\scriptsize InfinityStar}} &
\includegraphics[width=0.22\linewidth]{\vidframe{base}{0}{0}} &
\includegraphics[width=0.22\linewidth]{\vidframe{base}{0}{1}} &
\includegraphics[width=0.22\linewidth]{\vidframe{base}{0}{2}} &
\includegraphics[width=0.22\linewidth]{\vidframe{base}{0}{3}} &
\raisebox{5.2ex}{\rotatebox{270}{\tiny Seed~1}} \\
&
\includegraphics[width=0.22\linewidth]{\vidframe{base}{1}{0}} &
\includegraphics[width=0.22\linewidth]{\vidframe{base}{1}{1}} &
\includegraphics[width=0.22\linewidth]{\vidframe{base}{1}{2}} &
\includegraphics[width=0.22\linewidth]{\vidframe{base}{1}{3}} &
\raisebox{5.2ex}{\rotatebox{270}{\tiny Seed~2}} \\
&
\includegraphics[width=0.22\linewidth]{\vidframe{base}{2}{0}} &
\includegraphics[width=0.22\linewidth]{\vidframe{base}{2}{1}} &
\includegraphics[width=0.22\linewidth]{\vidframe{base}{2}{2}} &
\includegraphics[width=0.22\linewidth]{\vidframe{base}{2}{3}} &
\raisebox{5.2ex}{\rotatebox{270}{\tiny Seed~3}} \\
\multirow{3}{*}[3.5ex]{\rotatebox{90}{\scriptsize +Cond. Anneal.}} &
\includegraphics[width=0.22\linewidth]{\vidframe{cads}{0}{0}} &
\includegraphics[width=0.22\linewidth]{\vidframe{cads}{0}{1}} &
\includegraphics[width=0.22\linewidth]{\vidframe{cads}{0}{2}} &
\includegraphics[width=0.22\linewidth]{\vidframe{cads}{0}{3}} &
\raisebox{5.2ex}{\rotatebox{270}{\tiny Seed~1}} \\
&
\includegraphics[width=0.22\linewidth]{\vidframe{cads}{1}{0}} &
\includegraphics[width=0.22\linewidth]{\vidframe{cads}{1}{1}} &
\includegraphics[width=0.22\linewidth]{\vidframe{cads}{1}{2}} &
\includegraphics[width=0.22\linewidth]{\vidframe{cads}{1}{3}} &
\raisebox{5.2ex}{\rotatebox{270}{\tiny Seed~2}} \\
&
\includegraphics[width=0.22\linewidth]{\vidframe{cads}{2}{0}} &
\includegraphics[width=0.22\linewidth]{\vidframe{cads}{2}{1}} &
\includegraphics[width=0.22\linewidth]{\vidframe{cads}{2}{2}} &
\includegraphics[width=0.22\linewidth]{\vidframe{cads}{2}{3}} &
\raisebox{5.2ex}{\rotatebox{270}{\tiny Seed~3}} \\
\multirow{3}{*}[3.0ex]{\rotatebox{90}{\textbf{\scriptsize \methodname{}}}} &
\includegraphics[width=0.22\linewidth]{\vidframe{ours}{0}{0}} &
\includegraphics[width=0.22\linewidth]{\vidframe{ours}{0}{1}} &
\includegraphics[width=0.22\linewidth]{\vidframe{ours}{0}{2}} &
\includegraphics[width=0.22\linewidth]{\vidframe{ours}{0}{3}} &
\raisebox{5.2ex}{\rotatebox{270}{\tiny Seed~1}} \\
&
\includegraphics[width=0.22\linewidth]{\vidframe{ours}{1}{0}} &
\includegraphics[width=0.22\linewidth]{\vidframe{ours}{1}{1}} &
\includegraphics[width=0.22\linewidth]{\vidframe{ours}{1}{2}} &
\includegraphics[width=0.22\linewidth]{\vidframe{ours}{1}{3}} &
\raisebox{5.2ex}{\rotatebox{270}{\tiny Seed~2}} \\
&
\includegraphics[width=0.22\linewidth]{\vidframe{ours}{2}{0}} &
\includegraphics[width=0.22\linewidth]{\vidframe{ours}{2}{1}} &
\includegraphics[width=0.22\linewidth]{\vidframe{ours}{2}{2}} &
\includegraphics[width=0.22\linewidth]{\vidframe{ours}{2}{3}} &
\raisebox{5.2ex}{\rotatebox{270}{\tiny Seed~3}} \\
\end{tabular}
\vspace{2pt}
\parbox{0.98\linewidth}{\centering\scriptsize ``\emph{A nighttime scene transitions from a warm, golden moon to a cool, silvery moon.}''}
\end{minipage}
}
\begin{figure}[h!]
\centering
\setlength{\tabcolsep}{0.6pt}
\renewcommand{\arraystretch}{0.35}

\begin{minipage}{0.49\textwidth}
\centering
\newcommand{\vidframe}[3]{figures/results/video_samples/#0/0/##2_##3.jpg}
\newcommand{\vidprompt}{figures/results/video_samples/prompts/0.txt}
\begin{tabular}{>{\centering\arraybackslash}p{0.9em} cccc >{\centering\arraybackslash}p{0.5em}}
{} & \scriptsize Frame 1 & \scriptsize Frame 26 & \scriptsize Frame 51 & \scriptsize Frame 75 & {} \\
\multirow{3}{*}[1.5ex]{\rotatebox{90}{\scriptsize InfinityStar}} &
\includegraphics[width=0.22\linewidth]{\vidframe{base}{0}{0}} &
\includegraphics[width=0.22\linewidth]{\vidframe{base}{0}{1}} &
\includegraphics[width=0.22\linewidth]{\vidframe{base}{0}{2}} &
\includegraphics[width=0.22\linewidth]{\vidframe{base}{0}{3}} &
\raisebox{5.2ex}{\rotatebox{270}{\tiny Seed~1}} \\
&
\includegraphics[width=0.22\linewidth]{\vidframe{base}{1}{0}} &
\includegraphics[width=0.22\linewidth]{\vidframe{base}{1}{1}} &
\includegraphics[width=0.22\linewidth]{\vidframe{base}{1}{2}} &
\includegraphics[width=0.22\linewidth]{\vidframe{base}{1}{3}} &
\raisebox{5.2ex}{\rotatebox{270}{\tiny Seed~2}} \\
&
\includegraphics[width=0.22\linewidth]{\vidframe{base}{2}{0}} &
\includegraphics[width=0.22\linewidth]{\vidframe{base}{2}{1}} &
\includegraphics[width=0.22\linewidth]{\vidframe{base}{2}{2}} &
\includegraphics[width=0.22\linewidth]{\vidframe{base}{2}{3}} &
\raisebox{5.2ex}{\rotatebox{270}{\tiny Seed~3}} \\
\multirow{3}{*}[3.5ex]{\rotatebox{90}{\scriptsize +Cond. Anneal.}} &
\includegraphics[width=0.22\linewidth]{\vidframe{cads}{0}{0}} &
\includegraphics[width=0.22\linewidth]{\vidframe{cads}{0}{1}} &
\includegraphics[width=0.22\linewidth]{\vidframe{cads}{0}{2}} &
\includegraphics[width=0.22\linewidth]{\vidframe{cads}{0}{3}} &
\raisebox{5.2ex}{\rotatebox{270}{\tiny Seed~1}} \\
&
\includegraphics[width=0.22\linewidth]{\vidframe{cads}{1}{0}} &
\includegraphics[width=0.22\linewidth]{\vidframe{cads}{1}{1}} &
\includegraphics[width=0.22\linewidth]{\vidframe{cads}{1}{2}} &
\includegraphics[width=0.22\linewidth]{\vidframe{cads}{1}{3}} &
\raisebox{5.2ex}{\rotatebox{270}{\tiny Seed~2}} \\
&
\includegraphics[width=0.22\linewidth]{\vidframe{cads}{2}{0}} &
\includegraphics[width=0.22\linewidth]{\vidframe{cads}{2}{1}} &
\includegraphics[width=0.22\linewidth]{\vidframe{cads}{2}{2}} &
\includegraphics[width=0.22\linewidth]{\vidframe{cads}{2}{3}} &
\raisebox{5.2ex}{\rotatebox{270}{\tiny Seed~3}} \\
\multirow{3}{*}[3.0ex]{\rotatebox{90}{\textbf{\scriptsize \methodname{}}}} &
\includegraphics[width=0.22\linewidth]{\vidframe{ours}{0}{0}} &
\includegraphics[width=0.22\linewidth]{\vidframe{ours}{0}{1}} &
\includegraphics[width=0.22\linewidth]{\vidframe{ours}{0}{2}} &
\includegraphics[width=0.22\linewidth]{\vidframe{ours}{0}{3}} &
\raisebox{5.2ex}{\rotatebox{270}{\tiny Seed~1}} \\
&
\includegraphics[width=0.22\linewidth]{\vidframe{ours}{1}{0}} &
\includegraphics[width=0.22\linewidth]{\vidframe{ours}{1}{1}} &
\includegraphics[width=0.22\linewidth]{\vidframe{ours}{1}{2}} &
\includegraphics[width=0.22\linewidth]{\vidframe{ours}{1}{3}} &
\raisebox{5.2ex}{\rotatebox{270}{\tiny Seed~2}} \\
&
\includegraphics[width=0.22\linewidth]{\vidframe{ours}{2}{0}} &
\includegraphics[width=0.22\linewidth]{\vidframe{ours}{2}{1}} &
\includegraphics[width=0.22\linewidth]{\vidframe{ours}{2}{2}} &
\includegraphics[width=0.22\linewidth]{\vidframe{ours}{2}{3}} &
\raisebox{5.2ex}{\rotatebox{270}{\tiny Seed~3}} \\
\end{tabular}
\vspace{2pt}
\parbox{0.98\linewidth}{\centering\scriptsize ``\emph{\input{\vidprompt}}''}
\end{minipage}
\hfill%

\begin{minipage}{0.49\textwidth}
\centering
\newcommand{\vidframe}[3]{figures/results/video_samples/#1/1/##2_##3.jpg}
\newcommand{\vidprompt}{figures/results/video_samples/prompts/1.txt}
\begin{tabular}{>{\centering\arraybackslash}p{0.9em} cccc >{\centering\arraybackslash}p{0.5em}}
{} & \scriptsize Frame 1 & \scriptsize Frame 26 & \scriptsize Frame 51 & \scriptsize Frame 75 & {} \\
\multirow{3}{*}[1.5ex]{\rotatebox{90}{\scriptsize InfinityStar}} &
\includegraphics[width=0.22\linewidth]{\vidframe{base}{0}{0}} &
\includegraphics[width=0.22\linewidth]{\vidframe{base}{0}{1}} &
\includegraphics[width=0.22\linewidth]{\vidframe{base}{0}{2}} &
\includegraphics[width=0.22\linewidth]{\vidframe{base}{0}{3}} &
\raisebox{5.2ex}{\rotatebox{270}{\tiny Seed~1}} \\
&
\includegraphics[width=0.22\linewidth]{\vidframe{base}{1}{0}} &
\includegraphics[width=0.22\linewidth]{\vidframe{base}{1}{1}} &
\includegraphics[width=0.22\linewidth]{\vidframe{base}{1}{2}} &
\includegraphics[width=0.22\linewidth]{\vidframe{base}{1}{3}} &
\raisebox{5.2ex}{\rotatebox{270}{\tiny Seed~2}} \\
&
\includegraphics[width=0.22\linewidth]{\vidframe{base}{2}{0}} &
\includegraphics[width=0.22\linewidth]{\vidframe{base}{2}{1}} &
\includegraphics[width=0.22\linewidth]{\vidframe{base}{2}{2}} &
\includegraphics[width=0.22\linewidth]{\vidframe{base}{2}{3}} &
\raisebox{5.2ex}{\rotatebox{270}{\tiny Seed~3}} \\
\multirow{3}{*}[3.5ex]{\rotatebox{90}{\scriptsize +Cond. Anneal.}} &
\includegraphics[width=0.22\linewidth]{\vidframe{cads}{0}{0}} &
\includegraphics[width=0.22\linewidth]{\vidframe{cads}{0}{1}} &
\includegraphics[width=0.22\linewidth]{\vidframe{cads}{0}{2}} &
\includegraphics[width=0.22\linewidth]{\vidframe{cads}{0}{3}} &
\raisebox{5.2ex}{\rotatebox{270}{\tiny Seed~1}} \\
&
\includegraphics[width=0.22\linewidth]{\vidframe{cads}{1}{0}} &
\includegraphics[width=0.22\linewidth]{\vidframe{cads}{1}{1}} &
\includegraphics[width=0.22\linewidth]{\vidframe{cads}{1}{2}} &
\includegraphics[width=0.22\linewidth]{\vidframe{cads}{1}{3}} &
\raisebox{5.2ex}{\rotatebox{270}{\tiny Seed~2}} \\
&
\includegraphics[width=0.22\linewidth]{\vidframe{cads}{2}{0}} &
\includegraphics[width=0.22\linewidth]{\vidframe{cads}{2}{1}} &
\includegraphics[width=0.22\linewidth]{\vidframe{cads}{2}{2}} &
\includegraphics[width=0.22\linewidth]{\vidframe{cads}{2}{3}} &
\raisebox{5.2ex}{\rotatebox{270}{\tiny Seed~3}} \\
\multirow{3}{*}[3.0ex]{\rotatebox{90}{\textbf{\scriptsize \methodname{}}}} &
\includegraphics[width=0.22\linewidth]{\vidframe{ours}{0}{0}} &
\includegraphics[width=0.22\linewidth]{\vidframe{ours}{0}{1}} &
\includegraphics[width=0.22\linewidth]{\vidframe{ours}{0}{2}} &
\includegraphics[width=0.22\linewidth]{\vidframe{ours}{0}{3}} &
\raisebox{5.2ex}{\rotatebox{270}{\tiny Seed~1}} \\
&
\includegraphics[width=0.22\linewidth]{\vidframe{ours}{1}{0}} &
\includegraphics[width=0.22\linewidth]{\vidframe{ours}{1}{1}} &
\includegraphics[width=0.22\linewidth]{\vidframe{ours}{1}{2}} &
\includegraphics[width=0.22\linewidth]{\vidframe{ours}{1}{3}} &
\raisebox{5.2ex}{\rotatebox{270}{\tiny Seed~2}} \\
&
\includegraphics[width=0.22\linewidth]{\vidframe{ours}{2}{0}} &
\includegraphics[width=0.22\linewidth]{\vidframe{ours}{2}{1}} &
\includegraphics[width=0.22\linewidth]{\vidframe{ours}{2}{2}} &
\includegraphics[width=0.22\linewidth]{\vidframe{ours}{2}{3}} &
\raisebox{5.2ex}{\rotatebox{270}{\tiny Seed~3}} \\
\end{tabular}
\vspace{2pt}
\parbox{0.98\linewidth}{\centering\scriptsize ``\emph{\input{\vidprompt}}''}
\end{minipage}
\vspace{-0.7em}
\caption{\textbf{Additional Qualitative Comparisons on Video Generation (Cont'd).} InfinityStar~\cite{liu2025:infinitystar} produces videos with limited variation and diversity. Injecting noise into the text embeddings increases diversity, but at the cost of lower quality. Finally, applying our \textbf{scale-travel} refinement technique significantly improves the diversity of videos generated from the same prompt.}
\vspace{-\baselineskip}
\label{fig:video_qualitative_ap1}
\end{figure}
\begin{figure}[h!]
\centering
\setlength{\tabcolsep}{0.6pt}
\renewcommand{\arraystretch}{0.35}

\begin{minipage}{0.49\textwidth}
\centering
\newcommand{\vidframe}[3]{figures/results/video_samples/#2/2/##2_##3.jpg}
\newcommand{\vidprompt}{figures/results/video_samples/prompts/2.txt}
\begin{tabular}{>{\centering\arraybackslash}p{0.9em} cccc >{\centering\arraybackslash}p{0.5em}}
{} & \scriptsize Frame 1 & \scriptsize Frame 26 & \scriptsize Frame 51 & \scriptsize Frame 75 & {} \\
\multirow{3}{*}[1.5ex]{\rotatebox{90}{\scriptsize InfinityStar}} &
\includegraphics[width=0.22\linewidth]{\vidframe{base}{0}{0}} &
\includegraphics[width=0.22\linewidth]{\vidframe{base}{0}{1}} &
\includegraphics[width=0.22\linewidth]{\vidframe{base}{0}{2}} &
\includegraphics[width=0.22\linewidth]{\vidframe{base}{0}{3}} &
\raisebox{5.2ex}{\rotatebox{270}{\tiny Seed~1}} \\
&
\includegraphics[width=0.22\linewidth]{\vidframe{base}{1}{0}} &
\includegraphics[width=0.22\linewidth]{\vidframe{base}{1}{1}} &
\includegraphics[width=0.22\linewidth]{\vidframe{base}{1}{2}} &
\includegraphics[width=0.22\linewidth]{\vidframe{base}{1}{3}} &
\raisebox{5.2ex}{\rotatebox{270}{\tiny Seed~2}} \\
&
\includegraphics[width=0.22\linewidth]{\vidframe{base}{2}{0}} &
\includegraphics[width=0.22\linewidth]{\vidframe{base}{2}{1}} &
\includegraphics[width=0.22\linewidth]{\vidframe{base}{2}{2}} &
\includegraphics[width=0.22\linewidth]{\vidframe{base}{2}{3}} &
\raisebox{5.2ex}{\rotatebox{270}{\tiny Seed~3}} \\
\multirow{3}{*}[3.5ex]{\rotatebox{90}{\scriptsize +Cond. Anneal.}} &
\includegraphics[width=0.22\linewidth]{\vidframe{cads}{0}{0}} &
\includegraphics[width=0.22\linewidth]{\vidframe{cads}{0}{1}} &
\includegraphics[width=0.22\linewidth]{\vidframe{cads}{0}{2}} &
\includegraphics[width=0.22\linewidth]{\vidframe{cads}{0}{3}} &
\raisebox{5.2ex}{\rotatebox{270}{\tiny Seed~1}} \\
&
\includegraphics[width=0.22\linewidth]{\vidframe{cads}{1}{0}} &
\includegraphics[width=0.22\linewidth]{\vidframe{cads}{1}{1}} &
\includegraphics[width=0.22\linewidth]{\vidframe{cads}{1}{2}} &
\includegraphics[width=0.22\linewidth]{\vidframe{cads}{1}{3}} &
\raisebox{5.2ex}{\rotatebox{270}{\tiny Seed~2}} \\
&
\includegraphics[width=0.22\linewidth]{\vidframe{cads}{2}{0}} &
\includegraphics[width=0.22\linewidth]{\vidframe{cads}{2}{1}} &
\includegraphics[width=0.22\linewidth]{\vidframe{cads}{2}{2}} &
\includegraphics[width=0.22\linewidth]{\vidframe{cads}{2}{3}} &
\raisebox{5.2ex}{\rotatebox{270}{\tiny Seed~3}} \\
\multirow{3}{*}[3.0ex]{\rotatebox{90}{\textbf{\scriptsize \methodname{}}}} &
\includegraphics[width=0.22\linewidth]{\vidframe{ours}{0}{0}} &
\includegraphics[width=0.22\linewidth]{\vidframe{ours}{0}{1}} &
\includegraphics[width=0.22\linewidth]{\vidframe{ours}{0}{2}} &
\includegraphics[width=0.22\linewidth]{\vidframe{ours}{0}{3}} &
\raisebox{5.2ex}{\rotatebox{270}{\tiny Seed~1}} \\
&
\includegraphics[width=0.22\linewidth]{\vidframe{ours}{1}{0}} &
\includegraphics[width=0.22\linewidth]{\vidframe{ours}{1}{1}} &
\includegraphics[width=0.22\linewidth]{\vidframe{ours}{1}{2}} &
\includegraphics[width=0.22\linewidth]{\vidframe{ours}{1}{3}} &
\raisebox{5.2ex}{\rotatebox{270}{\tiny Seed~2}} \\
&
\includegraphics[width=0.22\linewidth]{\vidframe{ours}{2}{0}} &
\includegraphics[width=0.22\linewidth]{\vidframe{ours}{2}{1}} &
\includegraphics[width=0.22\linewidth]{\vidframe{ours}{2}{2}} &
\includegraphics[width=0.22\linewidth]{\vidframe{ours}{2}{3}} &
\raisebox{5.2ex}{\rotatebox{270}{\tiny Seed~3}} \\
\end{tabular}
\vspace{2pt}
\parbox{0.98\linewidth}{\centering\scriptsize ``\emph{\input{\vidprompt}}''}
\end{minipage}
\hfill%

\begin{minipage}{0.49\textwidth}
\centering
\newcommand{\vidframe}[3]{figures/results/video_samples/#3/3/##2_##3.jpg}
\newcommand{\vidprompt}{figures/results/video_samples/prompts/3.txt}
\begin{tabular}{>{\centering\arraybackslash}p{0.9em} cccc >{\centering\arraybackslash}p{0.5em}}
{} & \scriptsize Frame 1 & \scriptsize Frame 26 & \scriptsize Frame 51 & \scriptsize Frame 75 & {} \\
\multirow{3}{*}[1.5ex]{\rotatebox{90}{\scriptsize InfinityStar}} &
\includegraphics[width=0.22\linewidth]{\vidframe{base}{0}{0}} &
\includegraphics[width=0.22\linewidth]{\vidframe{base}{0}{1}} &
\includegraphics[width=0.22\linewidth]{\vidframe{base}{0}{2}} &
\includegraphics[width=0.22\linewidth]{\vidframe{base}{0}{3}} &
\raisebox{5.2ex}{\rotatebox{270}{\tiny Seed~1}} \\
&
\includegraphics[width=0.22\linewidth]{\vidframe{base}{1}{0}} &
\includegraphics[width=0.22\linewidth]{\vidframe{base}{1}{1}} &
\includegraphics[width=0.22\linewidth]{\vidframe{base}{1}{2}} &
\includegraphics[width=0.22\linewidth]{\vidframe{base}{1}{3}} &
\raisebox{5.2ex}{\rotatebox{270}{\tiny Seed~2}} \\
&
\includegraphics[width=0.22\linewidth]{\vidframe{base}{2}{0}} &
\includegraphics[width=0.22\linewidth]{\vidframe{base}{2}{1}} &
\includegraphics[width=0.22\linewidth]{\vidframe{base}{2}{2}} &
\includegraphics[width=0.22\linewidth]{\vidframe{base}{2}{3}} &
\raisebox{5.2ex}{\rotatebox{270}{\tiny Seed~3}} \\
\multirow{3}{*}[3.5ex]{\rotatebox{90}{\scriptsize +Cond. Anneal.}} &
\includegraphics[width=0.22\linewidth]{\vidframe{cads}{0}{0}} &
\includegraphics[width=0.22\linewidth]{\vidframe{cads}{0}{1}} &
\includegraphics[width=0.22\linewidth]{\vidframe{cads}{0}{2}} &
\includegraphics[width=0.22\linewidth]{\vidframe{cads}{0}{3}} &
\raisebox{5.2ex}{\rotatebox{270}{\tiny Seed~1}} \\
&
\includegraphics[width=0.22\linewidth]{\vidframe{cads}{1}{0}} &
\includegraphics[width=0.22\linewidth]{\vidframe{cads}{1}{1}} &
\includegraphics[width=0.22\linewidth]{\vidframe{cads}{1}{2}} &
\includegraphics[width=0.22\linewidth]{\vidframe{cads}{1}{3}} &
\raisebox{5.2ex}{\rotatebox{270}{\tiny Seed~2}} \\
&
\includegraphics[width=0.22\linewidth]{\vidframe{cads}{2}{0}} &
\includegraphics[width=0.22\linewidth]{\vidframe{cads}{2}{1}} &
\includegraphics[width=0.22\linewidth]{\vidframe{cads}{2}{2}} &
\includegraphics[width=0.22\linewidth]{\vidframe{cads}{2}{3}} &
\raisebox{5.2ex}{\rotatebox{270}{\tiny Seed~3}} \\
\multirow{3}{*}[3.0ex]{\rotatebox{90}{\textbf{\scriptsize \methodname{}}}} &
\includegraphics[width=0.22\linewidth]{\vidframe{ours}{0}{0}} &
\includegraphics[width=0.22\linewidth]{\vidframe{ours}{0}{1}} &
\includegraphics[width=0.22\linewidth]{\vidframe{ours}{0}{2}} &
\includegraphics[width=0.22\linewidth]{\vidframe{ours}{0}{3}} &
\raisebox{5.2ex}{\rotatebox{270}{\tiny Seed~1}} \\
&
\includegraphics[width=0.22\linewidth]{\vidframe{ours}{1}{0}} &
\includegraphics[width=0.22\linewidth]{\vidframe{ours}{1}{1}} &
\includegraphics[width=0.22\linewidth]{\vidframe{ours}{1}{2}} &
\includegraphics[width=0.22\linewidth]{\vidframe{ours}{1}{3}} &
\raisebox{5.2ex}{\rotatebox{270}{\tiny Seed~2}} \\
&
\includegraphics[width=0.22\linewidth]{\vidframe{ours}{2}{0}} &
\includegraphics[width=0.22\linewidth]{\vidframe{ours}{2}{1}} &
\includegraphics[width=0.22\linewidth]{\vidframe{ours}{2}{2}} &
\includegraphics[width=0.22\linewidth]{\vidframe{ours}{2}{3}} &
\raisebox{5.2ex}{\rotatebox{270}{\tiny Seed~3}} \\
\end{tabular}
\vspace{2pt}
\parbox{0.98\linewidth}{\centering\scriptsize ``\emph{\input{\vidprompt}}''}
\end{minipage}
\vspace{-0.7em}
\caption{\textbf{Additional Qualitative Comparisons on Video Generation (Cont'd).} InfinityStar~\cite{liu2025:infinitystar} produces videos with limited variation and diversity. Injecting noise into the text embeddings increases diversity, but at the cost of lower quality. Finally, applying our \textbf{scale-travel} refinement technique significantly improves the diversity of videos generated from the same prompt.}
\vspace{-\baselineskip}
\label{fig:video_qualitative_ap2}
\end{figure}

\end{document}